\documentclass[10pt,twocolumn,letterpaper]{article}

 \usepackage{iccv}              

\usepackage{amsfonts}
\usepackage{graphicx}
\usepackage{array}
\usepackage{booktabs} 
\usepackage{bm}
\usepackage{float}
\usepackage{times,amsmath,epsfig}
\usepackage{graphicx}
\usepackage{amssymb}
\usepackage{amsthm}
\usepackage[dvipsnames]{xcolor}
\usepackage{algorithm}
\usepackage{algorithmic}
\usepackage{multirow} 
\usepackage{array} 
\usepackage{booktabs} 
\usepackage{longtable}
\usepackage{bm}
\usepackage{verbatim} 
\usepackage{float}

\usepackage{color}
\usepackage{verbatim}
\usepackage{marvosym}
\usepackage{tcolorbox}
\newcolumntype{H}{>{\setbox0=\hbox\bgroup}c<{\egroup}@{}}
\usepackage{colortbl}
\usepackage{pifont}
\usepackage{makecell}

\definecolor{lightgray}{rgb}{0.9, 0.9, 0.9}
\usepackage{arydshln}
\setlist[itemize]{itemsep=0pt, parsep=0pt, partopsep=0pt, topsep=0pt}
\usepackage{tikz}     
\usepackage{xcolor}   
\newcommand{\textgray}[1]{\textcolor{gray}{#1}}
\definecolor{iccvblue}{rgb}{0.21,0.49,0.74}
\usepackage[pagebackref,breaklinks,colorlinks,allcolors=iccvblue]{hyperref}

\def\paperID{822} 
\def\confName{ICCV}
\def\confYear{2025}

\title{Semantics versus Identity: A Divide-and-Conquer Approach towards Adjustable Medical Image De-Identification}

\vspace{-8mm}
\author{\small   Yuan Tian$^{1}$
	~~
	Shuo Wang$^{2}$
	~~
	Rongzhao Zhang$^{1}$
	~~
	Zijian Chen$^{2,1}$
	~~
	Yankai Jiang$^{1}$\\
	\small	Chunyi Li$^{2,1}$
	~~
	Xiangyang Zhu$^{1}$
	~~
	Fang Yan$^{1}$
	~~
	Qiang Hu$^{3}$
	~~
	XiaoSong Wang$^{1}$
	~~
	Guangtao Zhai$^{2,1}${\textsuperscript{\Letter}}
	\\
	\small $^{1}$Shanghai AI Laboratory\\
	\small $^{2}$Institute of Image Communication and Network Engineering, Shanghai Jiao Tong Unversity \\
	\small $^{3}$Cooperative Medianet Innovation Center, Shanghai Jiao Tong Unversity \\	
}
\begin{document}
\maketitle

\begin{abstract}
	Medical\let\thefootnote\relax\footnote{\textsuperscript{\Letter}~Corresponding Author} imaging has significantly advanced computer-aided diagnosis, yet its re-identification (ReID) risks raise critical privacy concerns, calling for de-identification (DeID) techniques. Unfortunately, existing DeID methods neither particularly preserve medical semantics, nor are flexibly adjustable towards different privacy levels. To address these issues, we propose a divide-and-conquer framework comprising two steps: (1) \text{Identity-Blocking}, which blocks varying proportions of identity-related regions, to achieve different privacy levels; and (2) \text{Medical-Semantics-Compensation}, which leverages pre-trained Medical Foundation Models (MFMs) to extract medical semantic features to compensate the blocked regions. Moreover, recognizing that features from MFMs may still contain residual identity information, we introduce a \text{Minimum Description Length} principle-based feature decoupling strategy, to effectively decouple and discard such identity components.
	Extensive evaluations against existing approaches across seven datasets and three downstream tasks, demonstrates our state-of-the-art performance.
\end{abstract}

\vspace{-5mm}
\section{Introduction}
In the era of digital medicine, large-scale medical images, such as X-rays and fundus photographs~\cite{bernardes2011digital}, are routinely processed by AI-based diagnostic models~\cite{seyyed2021underdiagnosis,zhou2023foundation,fan2024deep,wang2021deep,zhou2022generalized} to aid clinical decision-making. However, the increasing availability of these images raises significant concerns regarding patient privacy~\cite{2013Protecting,cohen2019big,koch2013patient,2019Privacy}, calling for the research on medical image de-identification.

\begin{figure}[!thbp]
	\centering
	\tabcolsep=0.1mm
	\begin{tabular}{c}
		\includegraphics[width=0.99\linewidth]{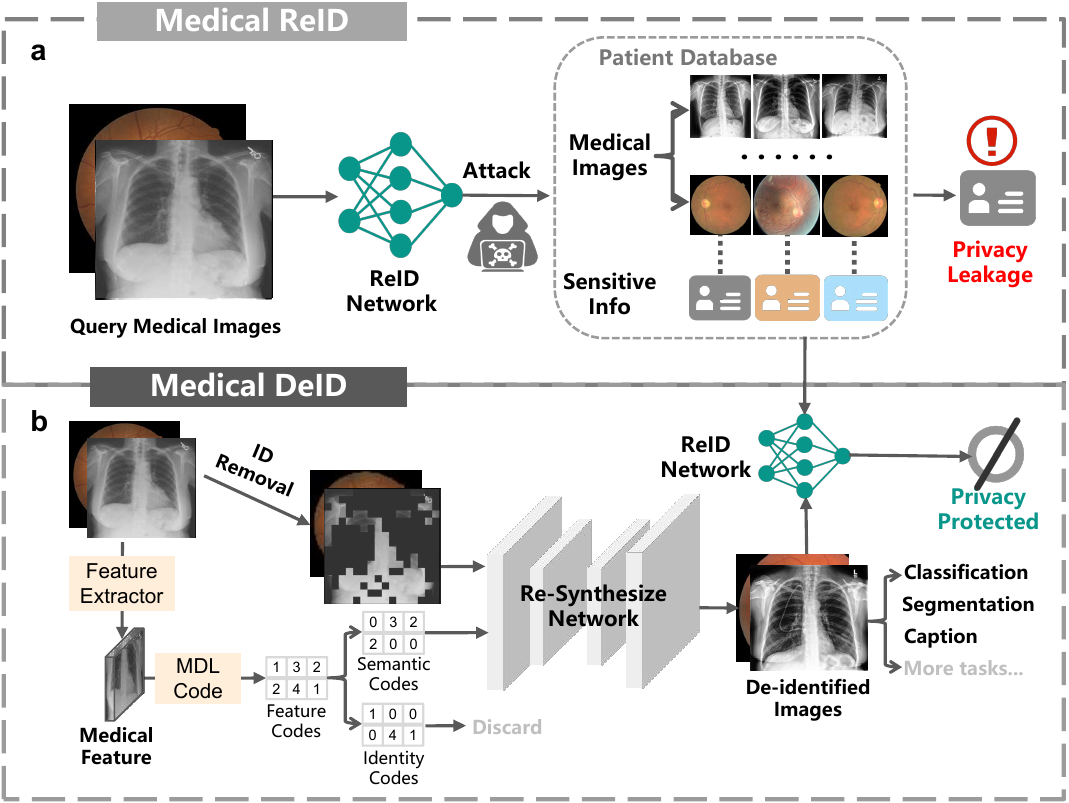}
	\end{tabular}
	\vspace{-3mm}
	\caption{(a) Given the query medical image, the ReID model can retrieve sensitive patient information from a leaked database. (b) Our DeID framework, removing identity and then compensating medical semantics, ensures adjustable identity protection, while preserving downstream task utility. Besides, a {Minimum Description Length} (MDL) principle-based code space is introduced, to decouple and discard the identity information in medical features.}
	\vspace{-6mm}
	\label{fig:task_description}
\end{figure}

Although explicit personal details such as patient name can be easily removed from medical image headers~\cite{2017A,aryanto2012implementation,rodriguez2010open} or burned-in texts~\cite{zhu2010automatic,tsui2012automatic}, re-identification (ReID) remains feasible for the intrinsic bio-identifiers, such as anatomical markers visible in chest X-rays~\cite{packhauser2022deep,1998Anatomy,tian2025towards}. This enables sensitive information breaches~\cite{bindschaedler2017tao,kaissis2020secure,tanner2017our}, compromising patient privacy (see Figure~\ref{fig:task_description}(a)).

Several studies have attempted to defend against ReID attacks. For instance, some approaches~\cite{fischl2012freesurfer,pydeface,hoopes2022synthstrip,cao2021personalized} focus on removing facial features to obfuscate identity. However, such methods cannot be applied to other body parts like the chest, where identity information is deeply interwoven with diagnostic semantics. Standard image filtering techniques, such as blurring~\cite{vishwamitra2017blur}, pixelation~\cite{hill2016effectiveness}, and masking~\cite{wang2022facemae}, indiscriminately degrade critical diagnostic details, thereby impairing downstream medical task performance. Moreover, under high privacy settings, the severe degradation of image quality further deteriorates task performance. Differential privacy methods~\cite{dwork2006differential,croft2021obfuscation,kumar2021novel,xue2021dp} mitigate identity information via noise injection, but this operation also perturbs diagnostic features. Identity adversarial learning methods~\cite{kim2021privacy,packhauser2023deep} train generators by jointly maximizing identity discrepancy between the generated and original images, while minimizing distortion of medical information. Nevertheless, given the inherent entanglement between identity and diagnostic features, these methods fail to preserve diagnostic semantics at high privacy levels adequately. Recently, diagnostic annotation-conditioned generative models~\cite{dumont2021overcoming,chen2024generative,hong20213d,wang2024semantic,tian2022fairness} have yielded promising results, yet they remain limited to task-specific semantics and cannot offer adjustable privacy levels. In summary, \textit{no existing method preserves task-generalizable semantics, while supporting a wide range of adjustable privacy levels}.

To address these issues, we introduce a novel divide-and-conquer framework DCM-DeID, which decouples identity removal from semantic preservation, to achieve semantic-rich yet adjustable de-identification.  Our approach includes three steps, i.e., \textit{ID-Blocking}, which masks identity-related regions to achieve adjustable privacy levels;
\textit{Medical Semantics Extraction}, which leverages pre-trained medical foundation models (MFMs)~\cite{zhang2024challenges,moor2023foundation} to extract semantic-rich medical features; \textit{Image Re-Synthesis}, which employs a diffusion model~\cite{rombach2022high,ho2020denoising} to synthesize de-identified images, given the above ID-masked image and the medical features. Moreover, considering that the features from MFMs may also contain some identity information, we introduce a novel minimum description length~\cite{grunwald2007minimum}-based feature decoupling strategy, which excludes identity-associated information from the vanilla MFM features in a minimum-codelength latent space. This effectively prevents the reintroduction of identity information during the image re-synthesis step.
Through the above designs, our approach achieves a better trade-off between DeID and diagnostic utilization than prior approaches.
Our contributions are:
\begin{itemize}
	\setlength{\itemsep}{0pt}
	\setlength{\parsep}{0pt}
	\setlength{\parskip}{0pt}
	\item We reveal that existing medical DeID methods fall short in preserving task-generalizable semantics, and do not adjust seamlessly across privacy levels.
	We build the first benchmark for this problem, by reproducing previous approaches fairly on seven datasets.
	\item We propose the DCM-DeID framework, which performs identity removal and medical semantics preservation in separate steps, enabling both adjustable privacy protection and medical task utility.
	\item We introduce a Minimum Description Length-based decoupling strategy, which decouples identity cues from medical features in a compact code space, further improving the privacy protection capability.
	\item Our framework demonstrates state-of-the-art performance. Extensive Analysis is also performed.
\end{itemize}

\section{Related Works}

\textbf{Image Privacy Protection.} Early methods applied low-level filters to obscure image details, including downsampling~\cite{dai2015towards}, blurring~\cite{vishwamitra2017blur}, and pixelation~\cite{hill2016effectiveness}. Later, encryption in alternate domains such as JPEG bitstreams~\cite{ra2013p3,tierney2013cryptagram} and DCT coefficients~\cite{yuan2015secure,yuan2015privacy} was explored, though these often introduced severe distortions that hindered downstream tasks. Homomorphic encryption~\cite{ziad2016cryptoimg,wang2017encrypted} addresses inference on encrypted images, but suffers from high computational cost~\cite{paillier1999public} and limited compatibility with advanced models like Vision Transformers~\cite{dosovitskiy2021an,tian2022ean,tian2023clsa}. Additionally, approaches for face images~\cite{maximov2020ciagan,gu2020password,cao2021personalized} leverage facial priors from StyleGAN~\cite{karras2019style} or face recognition networks~\cite{deng2019arcface,zhang2019adacos}, which may not readily generalize to other domains, such as the medical-domain images in our work.

\textbf{Medical Image De-Identification.} Early methods (e.g., FreeSurfer~\cite{fischl2012freesurfer}, PyDeface~\cite{pydeface}, SynthStrip~\cite{hoopes2022synthstrip}) focus on removing facial features in brain MRI. For common medical images, early approaches use pixel-domain filters (like blurring~\cite{vishwamitra2017blur} and pixelation~\cite{hill2016effectiveness}) or frequency-domain techniques~\cite{gaudio2023deepfixcx}, but these hand-crafted solutions also severely degrade the image details, leading to substantially degraded performance and visual quality~\cite{li2023agiqa,li2025perceptual,li2025information,aibench,chen2024gaia}. Differential Privacy methods~\cite{dwork2006differential,croft2021obfuscation,kumar2021novel,xue2021dp} inject noise into the training data, which compromises inference-time utility. More recent generative models~\cite{dumont2021overcoming,chen2024generative,hong20213d,wang2024semantic,tian2022fairness} synthesize images conditioned on disease labels or lesion masks. However, they tend to lack task generalizability and struggle to balance privacy-utility trade-offs, which are addressed by our approach.

\begin{figure*}[!thbp]
		\vspace{-5mm}
	\centering
	\tabcolsep=0.1mm
	\begin{tabular}{cc}
		\includegraphics[width=0.9 \linewidth]{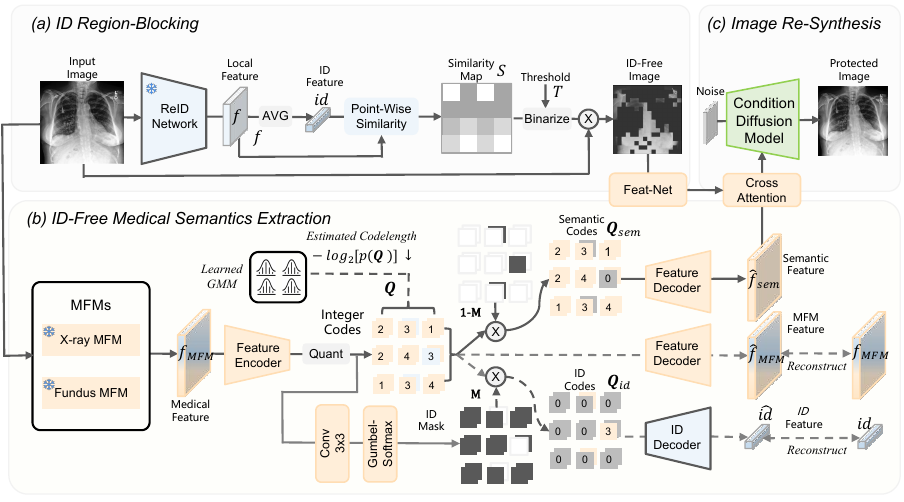}
	\end{tabular}
	\vspace{-4mm}
\caption{Overview of the proposed divide-and-conquer framework, DCM-DeID. (a) \textit{ID-Blocking}: A pre-trained ReID network produces the identity-similarity map, which is binarized by different thresholds to adjust privacy level. (b) \textit{ID-Free Medical Semantics Extraction}: Medical foundation models (MFMs) extract features that are encoded into a code space under the minimum-codelength regularization. A learned mask partitions the codes into identity- and medical semantics-related ones, where only the latter one is preserved. (c) \textit{Image Re-Synthesis}: A diffusion model re-synthesizes images that are privacy-preserving and semantics-rich, generalizing to various downstream tasks. We illustrate with X-ray images, but the framework is also applicable to other modalities such as fundus images. \includegraphics[height=0.7em]{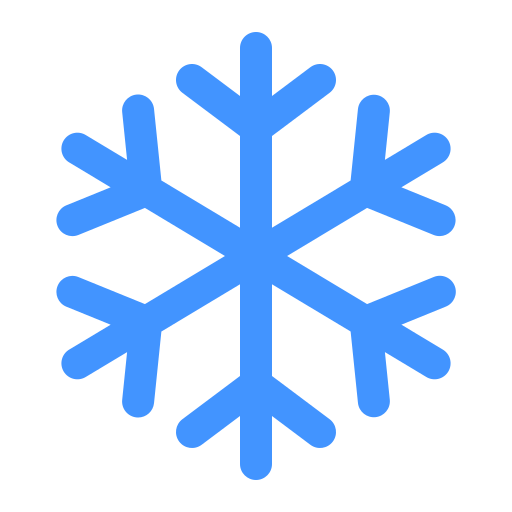} denotes frozen models, while gray dashed lines indicate components used solely for learning identity-semantic decoupling. The channel number of codes \(Q\) is arbitrary; two channels are shown for conciseness.
	$\otimes$ denotes the element-wise multiplication.
}
	\vspace{-5mm}
	\label{fig:method}
\end{figure*}

\textbf{Feature Decoupling.} Variational auto-encoder (VAE)-based works~\cite{higgins2017beta,shao2022rethinking} decouple representations, by constraining the variables in latent space independent. Generative adversarial network (GAN)-based methods~\cite{chen2016infogan,li2020mixnmatch} are unsupervised, leaving factors unaligned with explicit semantic or identity information. For face images, there are methods~\cite{deng2020disentangled,karras2019style,karras2020analyzing,tian2025medical} targeting identity separation.
However, these methods rely on strong facial priors that may not generalize to medical images. In contrast, motivated by the explicit entropy regularization terms adopted in data compression field~\cite{balle2016end,tian2024coding,hu20254dgc,hu2025varfvv,hu2025vrvvc,tian2023non,tian2021self,tian2024smc++,tian2024free,li2025image}, our approach effectively decouples identity in medical images, within a minimum-codelength space.

\vspace{-2mm}
\section{Methodology}
\vspace{-2mm}
In this section, we first describe the medical re-identification (ReID) models used for privacy attacks. Next, we introduce our de-identification model, which divides the task into two stages. First, identity information is removed via region blocking with an adjustable threshold. Second, lost medical semantics are compensated. This approach flexibly adjusts privacy while preserving rich, generalizable medical features for downstream tasks.

\subsection{Medical ReID Models}
Given a query medical image, ReID models aim to retrieve all images belonging to the same individual, from a medical record database.
Concretely, the model first extracts identity (ID) embedding from the query image, and then compare it with that of each image within the database.
Then, the image with the closest Euclidean distance is adopted as the re-identified image. 
We build two medical ReID models, i.e., ViT~\cite{dosovitskiy2021an} and VisionMamba~\cite{pmlrv235zhu24f}-based ones, which are separately adopted in the training and the evaluation stages.
These ReID models are optimized with a combination of classification loss and triplet loss~\cite{hermans2017defense}, following the previous object ReID work~\cite{he2021transreid}.

\subsection{A Divide-and-Conquer Approach}
To defend against attacks on medical ReID models, we propose DCM-DeID, a divide-and-conquer approach for medical image de-identification. DCM-DeID operates in three stages: \textit{ID Blocking}, which removes identity-related image regions; \textit{ID-Free Medical Semantics Extraction}, which extracts rich medical information without reintroducing identity information; and \textit{Image Re-Synthesis}, which generates the final de-identified medical image.

\textbf{ID-Blocking.}
Given an input image \(X \in \mathbb{R}^{3 \times H \times W}\), where $H$ and $W$ denote the image spatial scales, we use a ViT-based ReID model to extract local features \(f \in \mathbb{R}^{768 \times h \times w}\), where \(h = H / 16\) and \(w = W / 16\). Spatial average pooling is applied to \(f\) to obtain an identity embedding \(id \in \mathbb{R}^{768}\). For each spatial position in \(f\), the cosine similarity with \(id\) is computed, resulting in a similarity map \(S[i] = \cos(f[i], id)\), where $i$ denotes the spatial location.
Then, \(S\) is binarized by a threshold \(T\). Finally, the ID-blocked image is computed as:
$
X^{{noID}} = X \odot \operatorname{Upsample}(S > T),
$
where \(\operatorname{Upsample}\) denotes nearest-neighbor interpolation to match the resolution of \(S\) to \(X\).

\textbf{ID-Free Medical Semantics Extraction.} Although $X^{noID}$ effectively removes identity information, it inevitably distorts medical cues such as lung shadows. To amend this, we employ pre-trained medical foundation models (MFMs), e.g., MGCA~\cite{wang2022multi} for X-ray images, to extract rich medical feature $f_{{MFM}}$ from $X$. Since $f_{{MFM}}$ contains both semantic cues and local details that may encode identity, we introduce a feature decoupling strategy (Section~\ref{sec:feature_dis}) to decouple and remove the identity information, yielding the identity-free semantic feature $\hat{f}_{{sem}}$.

\textbf{Image Re-Synthesis.} Given $X^{noID}$ and $\hat{f}_{sem}$, a dual-conditioned diffusion model synthesizes the de-identified image that inherits the rich semantics within MFMs, while also protecting privacy. Since the synthesized image is highly realistic, it can be directly deployed to downstream medical AI applications, without further adaptation.
The model details are elaborated in Section~\ref{sec:diffusion_detail}.

\subsection{Medical Semantics Decoupling}
\label{sec:feature_dis}

Medical features extracted by the MFM encode both diagnostic semantics (e.g., lesion morphology) and identity-related cues (e.g., rib patterns in chest X-rays). For effective privacy-preserving, it is imperative to decouple these two types of information, and discard the identity cues. We achieve this by learning a minimum-length code space, and separating the two parts in this space.

\textbf{Theoretical Motivation.}
From an information-theoretic perspective, the Minimum Description Length (MDL) principle~\cite{grunwald2007minimum,barron1998minimum} states that the best representation for a given set of data is the one that minimizes the total codelength needed to describe the data, where each group of features tends to capture the independent or low-correlation information parts.
In our context, let \(\bm{Q}\) be the latent representation of the MFM feature $f_{\text{MFM}}$ and let \(H(\bm{Q})\) denote its expected codelength.
The MDL principle objective can be seen as balancing a reconstruction loss and a compression term, i.e., the so-called rate-distortion loss (RD loss)~\cite{berger2003rate}:
\vspace{-1mm}
\begin{equation}
	\mathcal{L}_{\text{code-all}} = \min_{\mathcal{E},\mathcal{D}} \quad \underbrace{\|{f}_{MFM} -  \hat{{f}}_{MFM}\|_2}_{\text{Feature Reconstruction}}+ \underbrace{\beta H(\bm{Q})}_{\text{Codelength}},
	\vspace{-1mm}
	\label{eq:bit_loss}
\end{equation}
where $\bm{Q} = \mathcal{E}(f_{MFM})$, \(\hat{{f}}_{MFM} = \mathcal{D}(\bm{Q})\), and $\beta$ denotes balancing weight. 
 \( \mathcal{E}\) and \(\mathcal{D}\) represent a pair of feature encoder and decoder networks.

\textbf{Discrete Code-based Codelength Estimation.}
Directly calculating the $H(\bm{Q})$ for the continuous variable $\bm{Q}$ is non-trivial~\cite{marsh2013introduction}.
Fortunately, the neural data compression community~\cite{balle2016end,minnen2018joint,balle2018variational,mentzer2018conditional} have verified that the codelength of \textit{integer} latent variables can be quite precisely estimated with a \textit{learnable entropy model}.
Therefore, we append the quantization operation at the tail of the encoder $\mathcal{E}$, to make elements within $\bm{Q}$ discrete values, and estimate its codelength.

Concretely, $\mathcal{E}$ comprises three residual blocks~\cite{he2016deep} with 256 channels, followed by a convolutional layer to reduce dimensionality and a rounding operation that outputs a 32-channel integer code $\bm{Q}$. The decoder network $\mathcal{D}$ is symmetric to $\mathcal{E}$, except it omits the rounding operation. During training, the straight-through estimator~\cite{mentzer2023finite} is employed to backpropagate gradients through the rounding step.

Following~\cite{balle2016end}, the expected codelength of encoding $\bm{Q}$ is calculated as the log-likelihood, i.e., $H(\bm{Q}) = -\log_2{p(\bm{Q})}$, where the probability $p(\bm{Q})$ is modeled using a Gaussian Mixture Model (GMM)~\cite{reynolds2009gaussian} with $K$ components:
\vspace{-2mm}
\begin{equation}
	p(\bm{Q}) = \sum_{k=1}^K w^k \cdot \mathcal{N}\left(\bm{Q}; \mu^k, e^{\sigma^k}\right),
	\vspace{-2mm}
\end{equation}
where $\{\mathbf{w}, \boldsymbol{\mu}, \boldsymbol{\sigma}\}$ are the learnable mixture weights, means, and log variance scalers of the GMM components, respectively, which are shared across spatial positions, not unshared along the channel axis~\cite{balle2016end}. Following~\cite{cheng2020learned}, $K$ is set to three. For each integer element $q \in \bm{Q}$, the probability is computed over the quantization bin~\cite{minnen2018joint,cover1999elements}, 
$
	p(q) = \mathcal{F}(q + 0.5) - \mathcal{F}(q - 0.5),
$
where
$
\mathcal{F}(x) = \sum_{k=1}^K w^k\, \Phi\Bigl(x; \mu^k, e^{\sigma^k}\Bigr)
$
is the cumulative distribution function (CDF) of the Gaussian Mixture Model (GMM),
$
	\Phi\Bigl(x; \mu, e^{\sigma}\Bigr) = \frac{1}{2}\left[ 1 + \operatorname{erf}\!\left( \frac{x-\mu}{\sqrt{2\, e^{\sigma}}} \right) \right].
$
We not that the CDF can be efficiently calculated by the modern deep learning framework such as PyTorch~\cite{paszke2019pytorch}.

\textbf{Learning of Identity-Associated Code Mask.}
A single convolution layer predicts a binary mask $\mathbf{M}$ from $\bm{Q}$, with the same dimensions as $\bm{Q}$. The Gumbel-Softmax algorithm~\cite{jang2016categorical} is applied to enable gradient propagation through the binary mask. The identity-associated codes are then obtained by element-wise masking,
$
	{\bm{Q}}_{\text{id}} = \bm{Q} \odot \mathbf{M}.
$
A lightweight convolutional network, composed of three residual blocks followed by average pooling, predicts the identity embedding $\hat{id}$ from ${\bm{Q}}_{\text{id}}$. Then, the RD loss for reconstructing identity can be given by:
\vspace{-1mm}
\begin{equation}
	\mathcal{L}_{\text{code-id}} =\|\hat{id} -id\|_2 +  \beta H({\bm{Q}}_{\text{id}}),
	\vspace{-1mm}
	\label{eq:id_loss}
\end{equation}
where $H(\tilde{\bm{Q}}_{\text{id}})$ is calculated similarly to $H(\bm{Q})$, sharing the same GMM parameters and balancing weight $\beta$ as in Equation~\ref{eq:bit_loss}, since they operate in the same latent space.

\textbf{Reconstruction of Medical Semantics.}
By suppressing identity-related codes via the inverse mask $(1-\mathbf{M})$, we obtain the semantics-part codes $\bm{Q}_{sem} = (1-\mathbf{M}) \otimes \bm{Q}$.
Finally, the final ID-free medical semantic feature is reconstructed as:
$
	\hat{f}_{{sem}} = \mathcal{D}(\bm{Q}_{sem}),
$
which preserves critical diagnostic semantics, excluding the identity information.

\subsection{Image Re-Synthesis Model}
\label{sec:diffusion_detail}
Given the ID-masked image \(X^{{noID}}\) and the ID-free medical semantic feature \(\hat{f}_{{sem}}\), we employ a diffusion model to synthesize de-identified medical images. 
First, we utilize a Feat-Net to project the high-resolution \(X^{{noID}}\) into the low-resolution feature \(f^{noID} \in \mathbb{R}^{512 \times \frac{H}{32} \times \frac{W}{32}}\). 
The Feat-Net consists of the VAE encoder from Stable Diffusion~\cite{rombach2022high}, followed by two convolution layers of kernel size 5 and stride size 2.
Next, we adopt a bi-directional cross-attention mechanism~\cite{chen2021crossvit} to fuse \(f^{{noID}}\) and \(\hat{f}_{{sem}}\), producing a fused feature \(f_{{fuse}} \in \mathbb{R}^{512 \times \frac{H}{32} \times \frac{W}{32}}\), which is further processed through a series of convolutional layers.
This produces a set of features with dimensions matching those of the UNet's intermediate feature maps within the diffusion model.
These features are added to the UNet layers, guiding the diffusion process toward two objectives: maintaining the privacy level of \(X^{{noID}}\), while preserving the medical semantics in \(\hat{f}_{sem}\).

\subsection{Learning Strategy}
The whole framework is end-to-end optimized, with the following objective,
\vspace{-2mm}
\begin{equation}
\mathcal{L}_{\text{total}} = 	\mathcal{L}_{\text{code-all}} + 	\mathcal{L}_{\text{code-id}} + 	\mathcal{L}_{\text{diffuse}},
\vspace{-2mm}
\end{equation}
where $\mathcal{L}_{\text{diffuse}}$ denotes the diffusion loss~\cite{ho2020denoising}.
We do not introduce the balancing weight, since we found directly adding the loss terms already achieves satisfactory results.

\begin{figure*}[!thbp] 
	\vspace{-5mm}
	\centering
	\newcommand{\widthscalefive}{0.25}
	\scriptsize
	\renewcommand{\arraystretch}{0.5}
	\tabcolsep = 1mm
	\scalebox{1}{
		\begin{tabular}{cccc}
			\includegraphics[width=0.24 \textwidth]{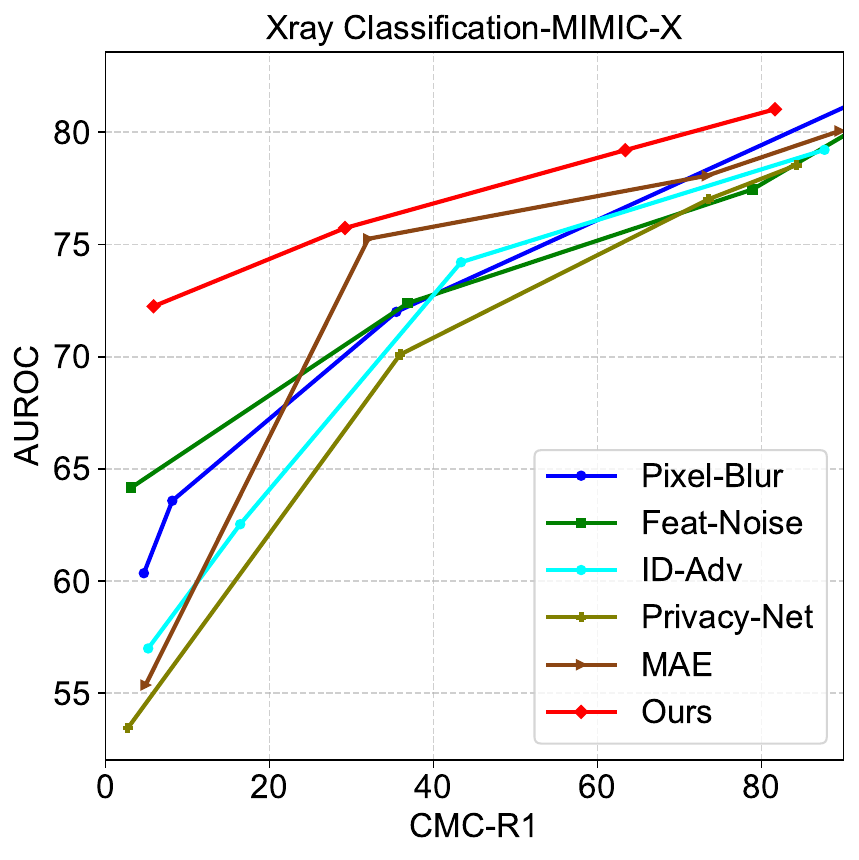}&
			\includegraphics[width=0.24 \textwidth]{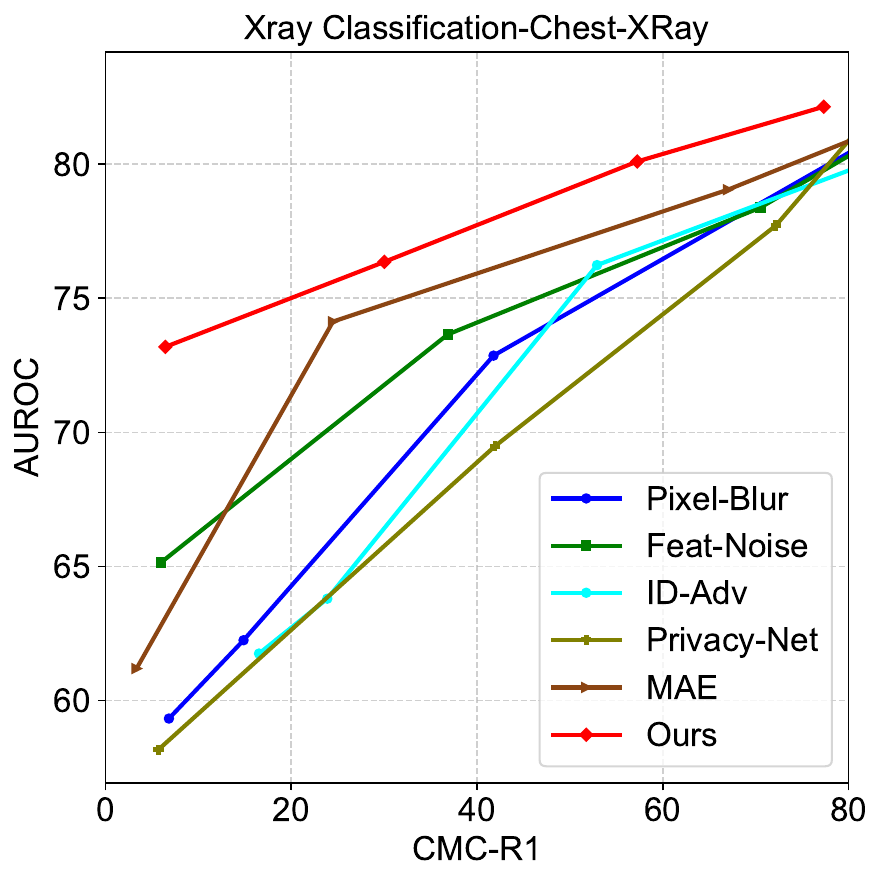} &
			\includegraphics[width=0.24 \textwidth]{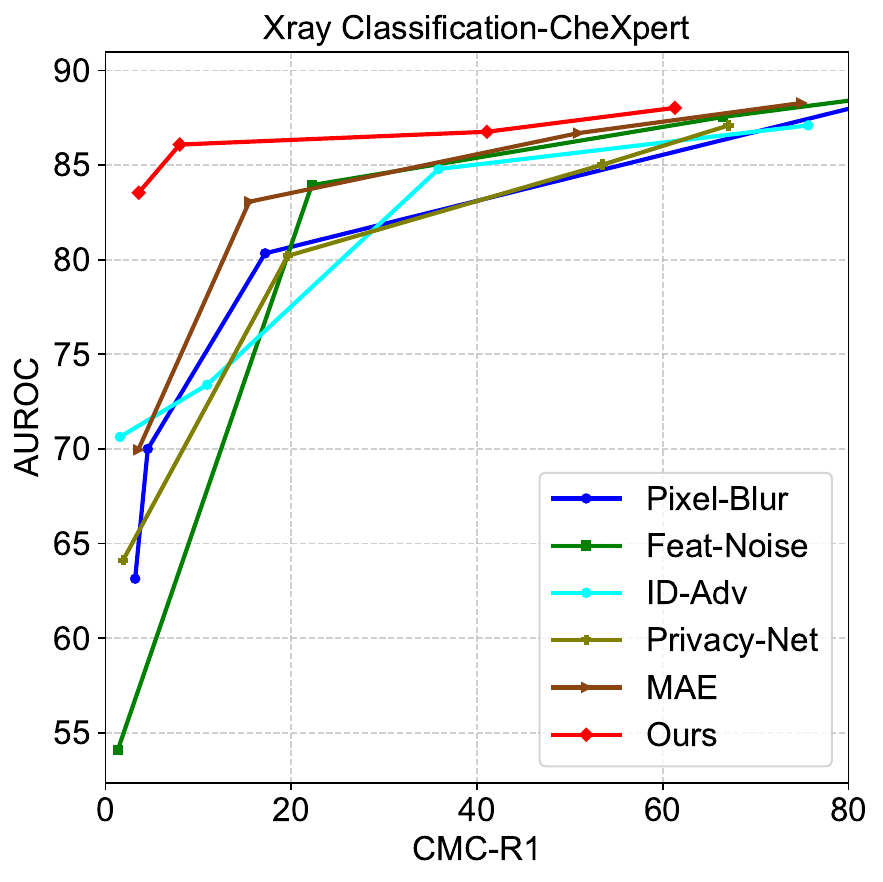}&
			\includegraphics[width=0.24 \textwidth]{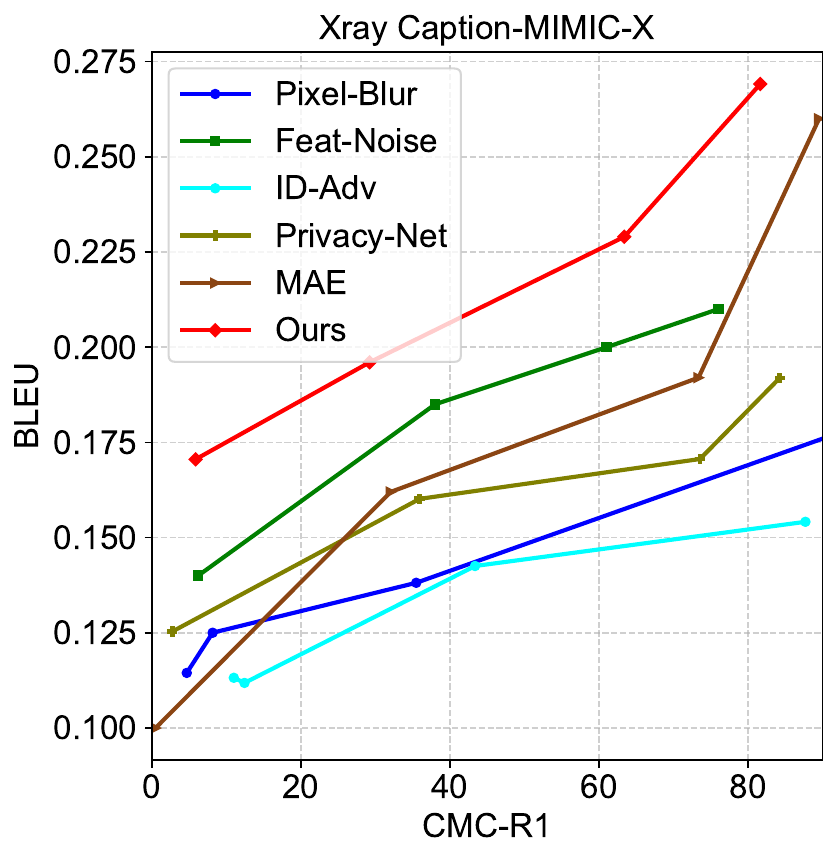}\\
			(a) & (b) & (c) &(d) \\
			\includegraphics[width=0.24 \textwidth]{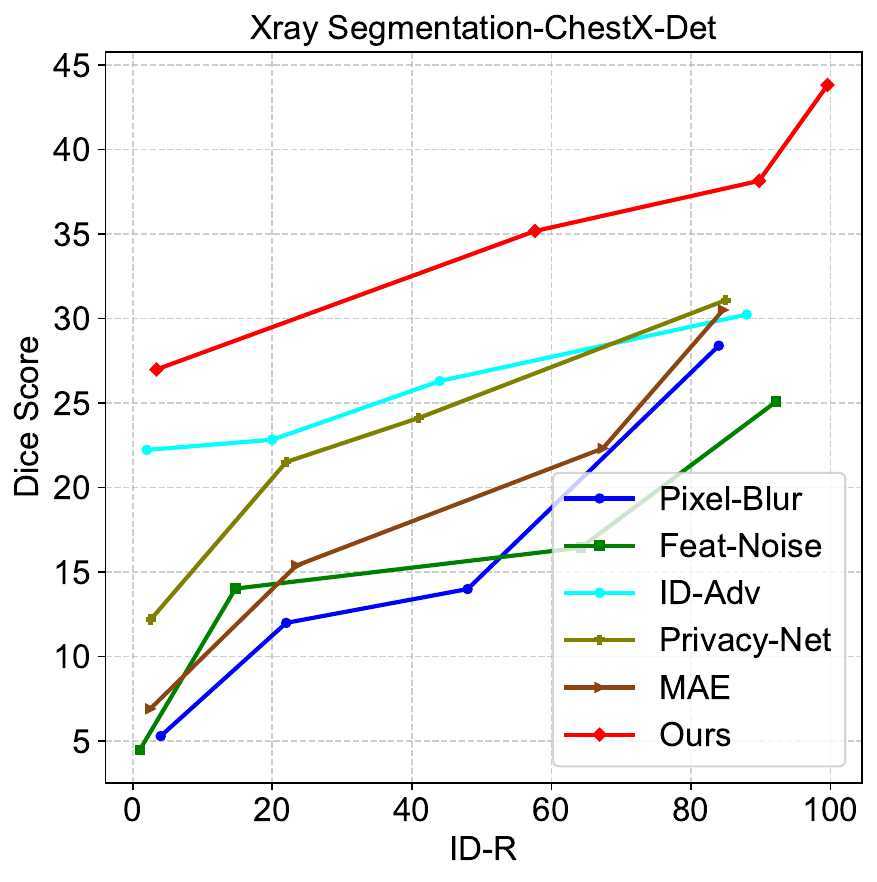} &
			\includegraphics[width=0.24 \textwidth]{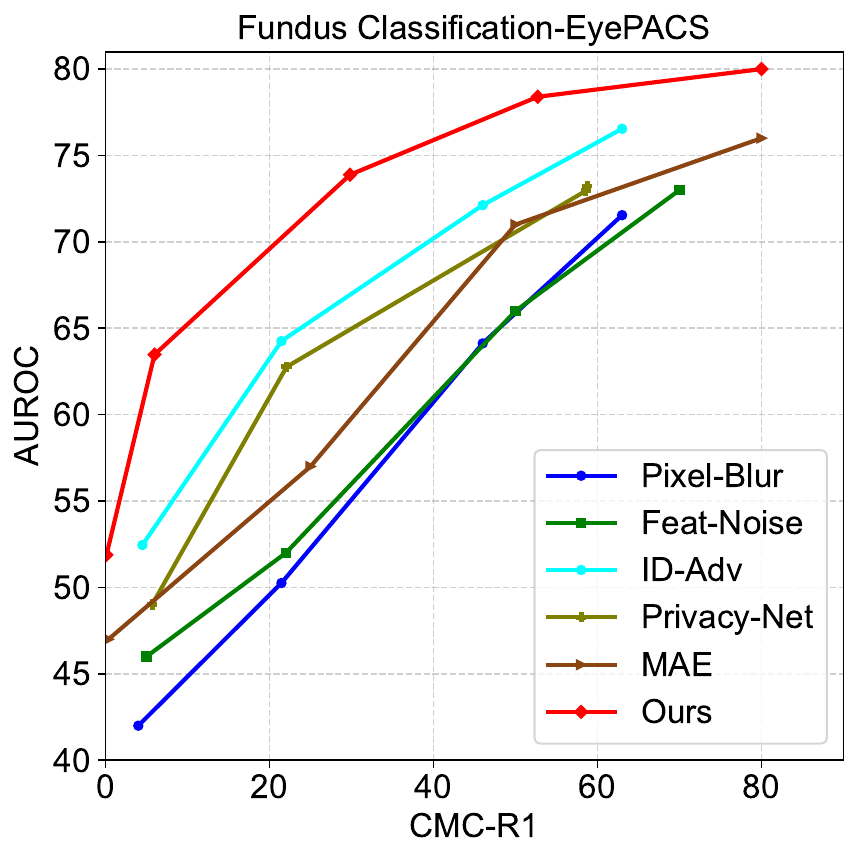}&
			\includegraphics[width=0.24 \textwidth]{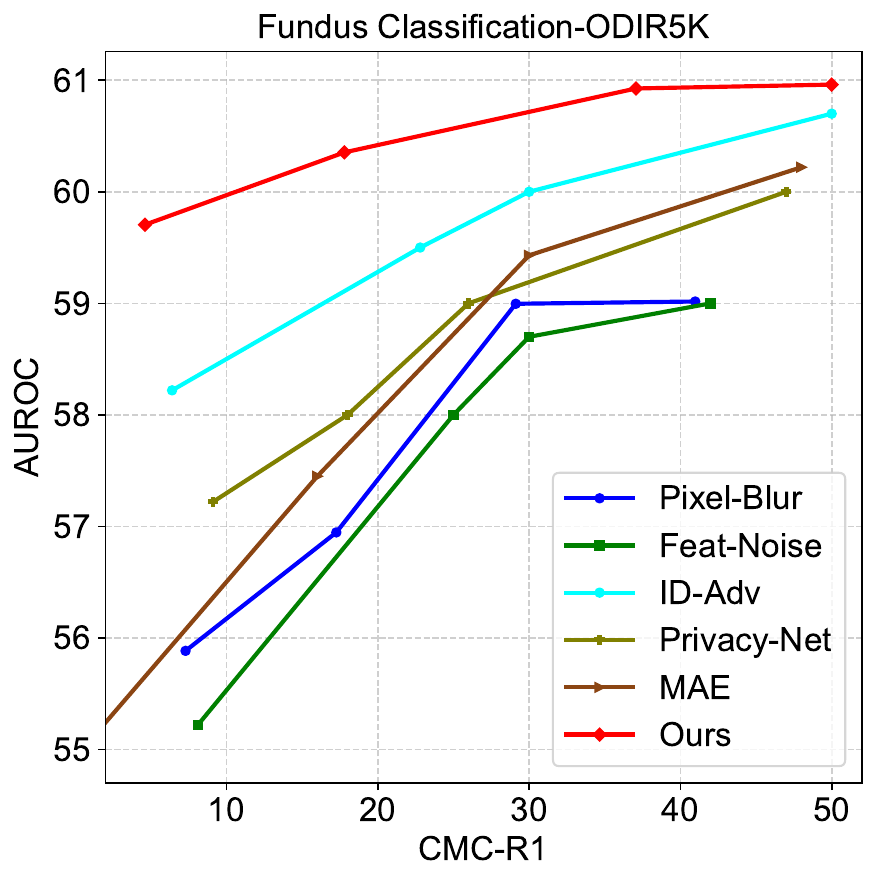}&
			\includegraphics[width=0.24 \textwidth]{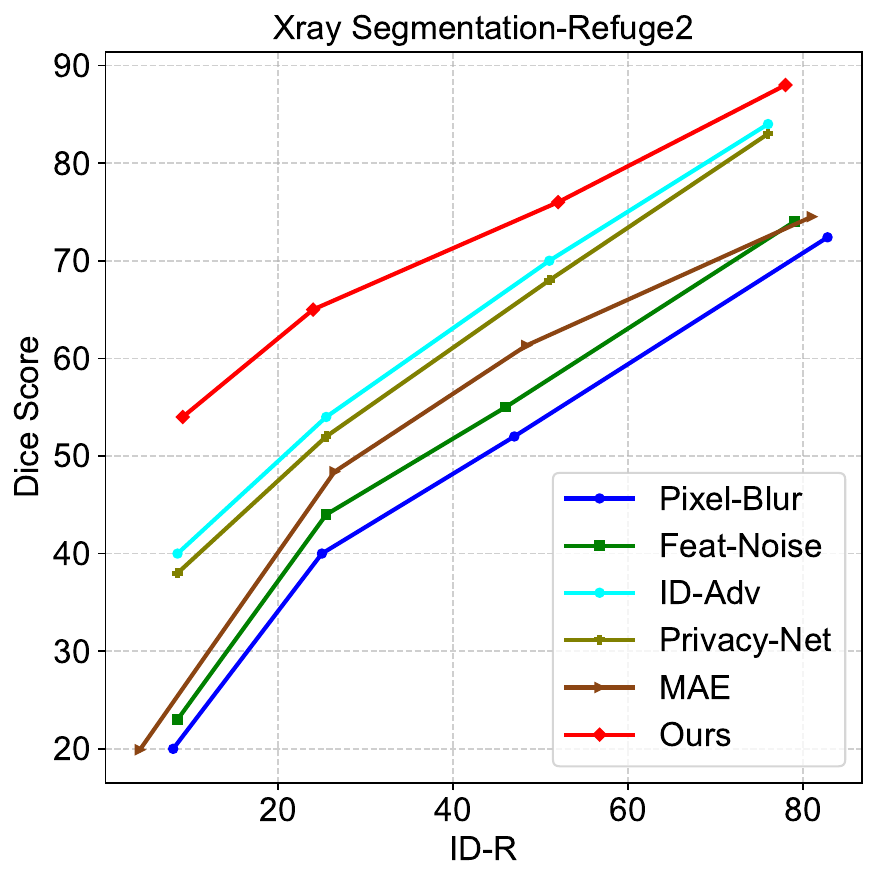}  \\
				(e) & (f) & (g) &(h) \\
		\end{tabular}
	}
	\vspace{-3mm}
	\caption{
		Identity-performance trade-off curves of various medical privacy protection methods.
	}
	\label{fig:performance}
		\vspace{-4mm}
\end{figure*}

\begin{table*}[htbp]
	\centering
	\renewcommand{\arraystretch}{0.5}
	\tabcolsep = 0.7mm
	\begin{tabular}{l l c c c c c c c c}
		\toprule
		\makecell{Attack\\ SR} & Method & \multicolumn{3}{c}{\makecell{X-ray Classify \\ AUROC (\%)}} & \multicolumn{1}{c}{\makecell{X-ray Caption \\ BLEU}} & \multicolumn{1}{c}{\makecell{X-ray Seg \\ Dice (\%)}} & \multicolumn{2}{c}{\makecell{Fundus Classify\\ AUROC (\%)}} & \multicolumn{1}{c}{\makecell{Fundus Seg \\ Dice (\%)}} \\
		\cmidrule(r){1-2} \cmidrule(r){3-5} \cmidrule(r){6-6} \cmidrule(r){7-7} \cmidrule(r){8-9} \cmidrule(r){10-10}
		& & MIMIC-X & Chest-XRay & CheXpert & MIMIC-X & ChestX-Det & EyePACS & ODIR5K & Refuge2 \\
		\midrule
		\multirow{6}{*}{10\%} 
		& Pixel-Blur~\cite{vishwamitra2017blur} & 64.15 & 60.48 & 74.45 & 0.1259 & 7.53  & 44.83 & 56.17 & 22.35 \\
		& Feat-Noise~\cite{zhai2018autoencoder} & 65.84 & 66.25 & 66.49 & 0.1453 & 10.74 & 47.76 & 55.53 & 24.85 \\
		& ID-Adv~\cite{packhauser2023deep}      & 59.36 & 61.75 & 73.11 & 0.1131 & 22.23 & 56.28 & 58.50 & 41.23 \\
		& Privacy-Net~\cite{kim2021privacy}     & 57.09 & 59.51 & 71.47 & 0.1329 & 15.75 & 52.63 & 57.29 & 39.23 \\
		& MAE~\cite{he2022masked}               & 59.11 & 65.26 & 77.05 & 0.1186 & 9.94  & 50.91 & 56.50 & 27.33 \\
		& Ours        & \textbf{72.86} & \textbf{73.66} & \textbf{86.12} & \textbf{0.1750} & \textbf{27.98} & \textbf{65.23} & \textbf{59.96} & \textbf{54.66} \\
		\midrule
		\multirow{6}{*}{20\%} 
		& Pixel-Blur~\cite{vishwamitra2017blur} & 67.23 & 64.27 & 80.67 & 0.1307 & 11.25  & 49.56 & 57.42 & 34.11 \\
		& Feat-Noise~\cite{zhai2018autoencoder} & 68.27 & 69.01 & 80.78 & 0.1595 & 14.28 & 51.29 & 57.17 & 37.20 \\
		& ID-Adv~\cite{packhauser2023deep}      & 64.08 & 62.71 & 77.53 & 0.1193 & 22.83 & 63.24 & 59.28 & 49.47 \\
		& Privacy-Net~\cite{kim2021privacy}     & 62.11 & 62.63 & 80.25 & 0.1434 & 20.55 &61.01 & 58.25& 47.47 \\
		& MAE~\cite{he2022masked}               & 66.43 & 71.38 & 83.52 & 0.1383 & 13.99  & 54.97 & 58.01 & 40.13 \\
		& Ours                            & \textbf{74.35} & \textbf{75.01} & \textbf{86.32} & \textbf{0.1859} & \textbf{29.49} & \textbf{69.59} & \textbf{60.41} & \textbf{62.04} \\
		\midrule
		& \textgray{Original}           & \textgray{82.13} & \textgray{84.82} & \textgray{87.24} & \textgray{0.3218} & \textgray{52.89} & \textgray{81.46} & \textgray{61.53} & \textgray{90.08} \\
		\bottomrule
	\end{tabular}
	\vspace{-2mm}
	\caption{
		Performance comparison of medical image privacy protection methods, under different attack Success-Rates (SR).
		For measuring SR, we adopt CMC-R1 metric for MIMIC-X, Chest-Xray, CheXpert, EyePACS, and ODIR5K, using ID-R metric for ChestX-Det and Refuge2.
		\textgray{Original} denotes the performance on original images, which is the performance upper-bound of privacy-removal images.
	}
	\vspace{-4mm}
	\label{tab:performance}
\end{table*}

\vspace{-2mm}
\section{Experiments}
\subsection{Implementation Details}
For \textit{Med-ReID} models, we adopt the AdamW optimizer~\cite{loshchilov2017decoupled} during training, with a learning rate of 1e-5 scheduled by cosine decaying strategy and a weight decay of 1e-2. The training process consists of 300,000 steps. The batch size is 256.
We apply random cropping and blurring as image augmentation strategies, and the input image resolution to networks is \(256 \times 256\). The ViT-based models are initialized with CLIP-pretrained weights~\cite{radford2021learning}, while the VisionMamba-based models are initialized with ImageNet-pretrained weights~\cite{deng2009imagenet}.
Training a single ReID model takes about 24 hours with four NVIDIA RTX 4090 GPUs.

For \textit{DCM-DeID} model, the UNet within the diffusion model follows the same architecture as the Stable Diffusion~\cite{rombach2022high}, also performing the diffusion procedure in the latent space. The feature channels within UNet are reduced to [128, 256, 512, 1024], for the four stages of both the down-pathway and up-pathway, to reduce computational cost.
The identity-similarity map threshold $T$ is defined as the \(r\)-th quantile of the similarity map \(S\).
\(r\) is selected from [0.95, 0.7, 0.4, 0.2] to cover wide privacy levels.
We adopt MGCA-ResNet~\cite{wang2022multi} and RetFound-ViT~\cite{zhou2023foundation} MFMs for X-ray and fundus images, respectively.
During training, we apply random flipping and random cropping 256 $\times$ 256 patches for data augmentation.
The codelength loss term weight $\beta$ is set to 0.5.
At test time, we resize the shorter side of the images to 256 and then center-crop the middle 256$\times$256 region.
The learning rate is set to 1e-4 and is gradually decayed with the cosine annealing strategy~\cite{loshchilov2016sgdr}. The total number of training steps is 800,000.
The mini-batch size is 64.
We utilize the AdamW optimizer~\cite{loshchilov2017decoupled} implemented in PyTorch~\cite{paszke2019pytorch} with CUDA support. The momentum parameters are set as $\beta_1 = 0.9$ and $\beta_2 = 0.99$, and the gradient norm is clipped to a maximum value of 1.
The entire training process takes about three days on a machine equipped with eight NVIDIA RTX 4090 GPUs.

\subsection{Datasets}
We evaluate our approach on two medical image modalities: chest X-rays and eye fundus photographs, with {seven} public datasets. For the \textit{chest X-ray} modality, we split the {MIMIC-X} dataset~\cite{johnson2019mimic} into training, validation, and test sets using an 8:1:1 ratio. For the {Chest-Xray} and {CheXpert} datasets, we randomly select 10\% patients as the test set. We also adopt the ChestX-Det dataset~\cite{lian2021structure} to evaluate the X-ray segmentation task.
For the \textit{eye fundus} modality, we divide the {EyePACS} dataset~\cite{diabetic-retinopathy-detection} into training, validation, and test sets with an 8:1:1 ratio, and we use the Refuge2 dataset~\cite{fang2022refuge2} to evaluate the fundus segmentation task. ODIR5K~\cite{bhati2023discriminative} is also adopted for evaluating the fundus classification task of systemic diseases such as hypertension.
Note that only {MIMIC-X} and {EyePACS} are used during training; all other datasets, which differ in environment, demographics, and imaging devices, are never seen during training, to evaluate the domain generalizability of our approach. MIMIC-X, CheXpert, Chest-Xray, EyePACS, ChestXDet, and ODIR5K contain 377K, 224K, 112K, 88K, 3.6K, and 5K
images.

\subsection{Reproduced Privacy Protection Methods}
We implement several privacy protection methods, comparing them with our approach in a fair setting.

\noindent \textbf{Pixel-Blur}~\cite{vishwamitra2017blur}.
This method applies a Gaussian blur to the input image. We experiment with standard deviations of $\{1, 5, 10, 20\}$ to vary the level of de-identification.

\noindent \textbf{Feat-Noise}~\cite{zhai2018autoencoder}.
We train an autoencoder~\cite{zhai2018autoencoder} and inject Gaussian noise into its latent features. The noise level is selected from $\{0.1, 0.8, 0.85, 0.9, 0.95\}$.

\noindent \textbf{ID-Adv}~\cite{packhauser2023deep}. 
A UNet is trained to generate a de-identified image $Y$ from the original image $X$, optimizing the loss
$
	\mathcal{L} = \lambda\, \cos({id}_{X},{id}_{Y}) + \|{med}_{X} - {med}_{Y}\|_2 + \mathcal{L}_{reg},
$
where ${id}_{X}$ and ${id}_{Y}$ are identity features extracted by a ViT-based ReID model, and ${med}_{X}$ and ${med}_{Y}$ are medical features obtained from MFMs same as our approach. $\mathcal{L}_{reg}$ is a GAN regularization loss ensuring visual plausibility, $\cos(\cdot,\cdot)$ denotes cosine similarity, and $\|\cdot\|_2$ the $\ell_2$ norm. The trade-off weight $\lambda$ is chosen from $\{0.1, 0.5, 1, 2\}$.

\noindent \textbf{Privacy-Net}~\cite{kim2021privacy}.  
This method updates the identity model and the de-identification network adversarially, enhancing de-identification performance. The original Privacy-Net focuses solely on segmentation tasks, supervised by segmentation masks. To enable task-agnostic de-identification, we train it using the same objective as ID-Adv. Since the identity model is adversarially updated and are stronger, we use smaller $\lambda$ values compared to ID-Adv, i.e., $\{0.05, 0.25, 0.5, 1\}$.

\noindent \textbf{MAE}~\cite{he2022masked}.  
Following~\cite{wang2022facemae}, we transfer the concept of masked auto-encoders (MAE)~\cite{he2022masked} to the adjustable privacy protection problem, by masking a random proportion of patches to obscure identity information.
It adopts the same diffusion model as our approach to generate the masked regions.
This model can also serve as a degenerated version of our model, where both semantic compensation and identity-region similarity designs are removed.

\subsection{Downstream Task Models}
For the \textit{identity recognition}, we adopt the VisionMamba-based ReID model, which differs from the ViT-based model employed during the training of privacy protection methods, ensuring the method generalization capability across different ReID models.
For the \textit{X-ray classification}, we use the ViT model pre-trained with Med-UniC~\cite{wan2024med}. For \textit{X-ray captioning}, we employ the visual-language model CXR-LLaVA-v2~\cite{lee2025cxr}, which is specifically designed for X-ray images. For \textit{X-ray segmentation}, we adopt CGRSeg~\cite{ni2024context}. For \textit{fundus classification}, we use the ViT model pre-trained with KeepFit~\cite{wu2024mm}. Finally, for \textit{fundus segmentation}, given the limited dataset size, we employ nnUNet~\cite{isensee2021nnu}.

\subsection{Evaluation Metrics}
For privacy evaluation, we adopt the cumulative matching characteristics (CMC)~\cite{bolle2005relation} at Rank-1, i.e., CMC-R1, on datasets with patient ID information available (i.e., MIMIC-X, Chest-Xray, CheXpert, EyePacs, and ODIR5K). For datasets without patient ID information (i.e., CheX-det and REFUGE2), we adopt the recognition rate, i.e., ID-R, which determines whether the distance between the ID feature of the original and de-identified image exceeds a predefined threshold. 
The thresholds are set to 1.1 and 1.35 for the X-ray and fundus modalities, respectively, based on statistics from the validation sets of MIMIC-X and EyePACS.
For the disease diagnosis task, we employ the area under the receiver operating characteristic curve (AUROC) metric~\cite{bradley1997use}; for the image captioning task, we use the bilingual evaluation understudy (BLEU) metric~\cite{papineni2002bleu}; and for the image segmentation task, we adopt the Dice score metric~\cite{bertels2019optimizing}.

\subsection{Results}

\textbf{X-ray Classification.}
As shown in Table~\ref{tab:performance}, our method substantially outperforms other approaches, achieving AUROCs of 72.86\%, 73.66\%, and 86.12\% on MIMIC-X, Chest-XRay, and Chexpert, respectively, under CMC-R1=10\%. Notably, although our model is trained on MIMIC-X, it generalizes well to the other two datasets.

Among the other compared approaches, Feat-Noise obtains the second-best performance, i.e., AUROC of 65.84\% at CMC-R1=10\% on MIMIX-X, by condensing image pixels into a compact latent feature space. In contrast, methods that jointly optimize a trade-off between de-identification and medical preservation, i.e., ID-Adv and PrivacyNet, yield unsatisfactory performances. As shown in Figure~\ref{fig:performance} (a), under the ID-R=5\% setting, ID-Adv and PrivacyNet attain AUROCs of only 56.32\% and 54.21\% on MIMIC-X, respectively, which are much lower than the simple pixel blurring baseline (61.62\%). This indicates that directly optimizing the two conflicting objectives is suboptimal.
In contrast, our approach decouples the objectives into two separate steps, identity removal and medical semantic compensation, achieving consistently superior performance.

As for MAE, which employs the same diffusion model as ours, it achieves competitive results at a high attacking rate, with an AUROC of 76.12\% @CMC-R1=40\% on MIMIC-X, outperforming all other approaches except ours. 
However, at a low attacking rate CMC-R1=10\%, it falls behind our method by over 13\% AUROC. This highlights that our superior performance is not solely due to the generative power of the diffusion model, but rather stems from the effectiveness of our core idea of semantic compensation.

\textbf{X-ray Caption.}
As shown in Table~\ref{tab:performance}, our method attains a BLEU score of 0.1750, remarkably surpassing Pixel-Blur (0.1259), Feat-Noise (0.1453), and MAE (0.1186), at CMC-R1=10\%. This proves that our approach can comprehensively preserve the clinic-required information, beyond only the classification label.

\textbf{X-ray Segmentation.}
Furthermore, we evaluate the methods on a fine-grained task: segmentation. As shown in Table~\ref{tab:performance}, at ID-R1=10\%, our method achieves a Dice score of 27.98\%, outperforming Pixel-Blur (7.53\%), Feat-Noise (10.74\%), Privacy-Net (15.75\%), ID-Adv (22.23\%), and MAE(9.94\%). 
This proves that our semantic compensation scheme not only preserves the global semantics for classification, but also effectively retains the local semantics for segmentation. Pixel-Blur and Feat-Noise perform poorly, since they severely corrupts the image details. In contrast, ID-Adv and Privacy-Net, which incorporate a medical feature-matching loss, achieve slightly decent performance, but still lag far behind our approach. 
For instance, under ID-R=80\%, our method outperforms Privacy-Net by approximately 8\%, as shown in Figure~\ref{fig:performance}(e).

\textbf{Fundus Classification.}
Beyond X-ray images, our method also proves effective on fundus data. For instance, on EyePACS and ORID5K, our approach outperforms the second-best competitor ID-Adv by about 9\% and 1\%, respectively. These results confirm that our approach generalizes well across different imaging modalities.

\textbf{Fundus Segmentation.}
Our method achieves a Dice score of 54.66\% on REFUGE2 at ID-R1=10\% , largely surpassing MAE (27.33\%), Privacy-Net (39.23\%), and ID-Adv (41.23\%).
This further validates that our method also effectively preserves fine-grained semantic cues of eye fundus.

\begin{figure}[!tbp] 
	\centering
	\small
	\newcommand{\widthscalefive}{0.25}
	\tabcolsep = 0.5mm
	\scalebox{1}{
		\begin{tabular}{cc}
			\includegraphics[width=0.48 \linewidth]{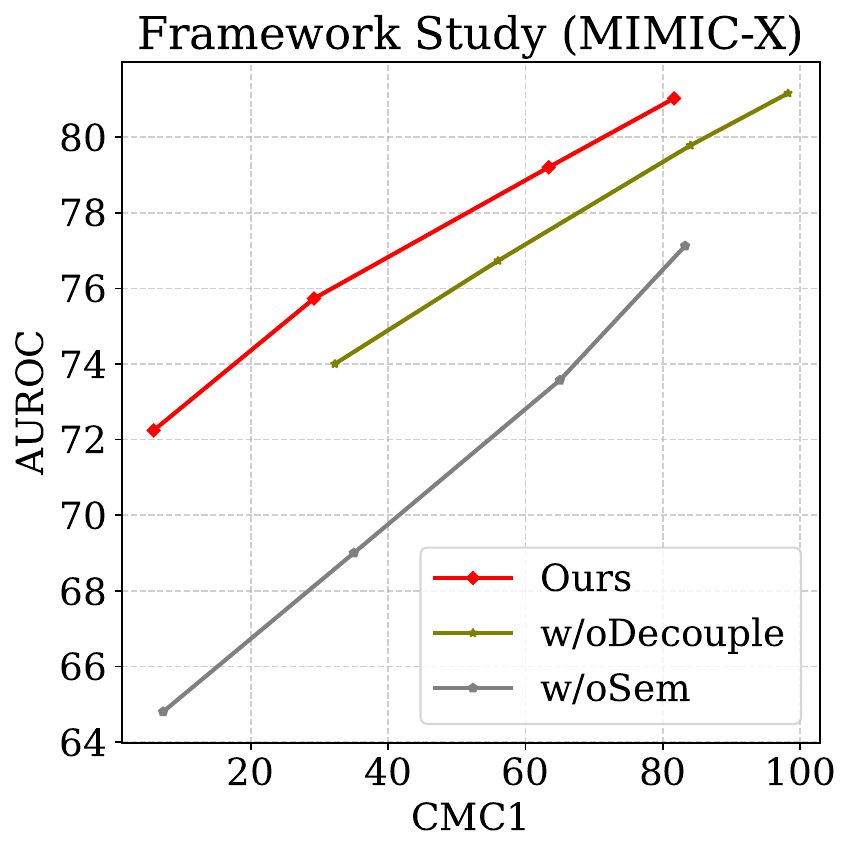}&
			\includegraphics[width=0.48 \linewidth]{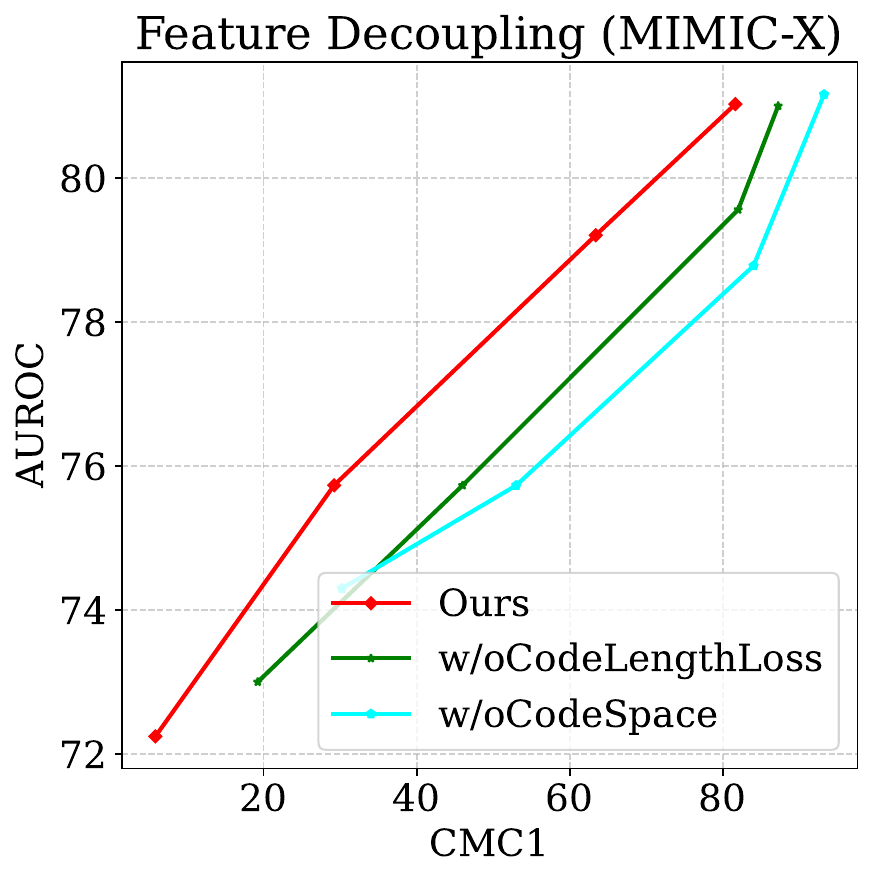}  \\
		\end{tabular}
	}
	\vspace{-4mm}
	\caption{
		(\textit{Left}) Ablation on the framework design.
		(\textit{Right}) Ablation study on the feature decoupling strategy.
	}
	\vspace{-4mm}
	\label{fig:ablation}
\end{figure}
\begin{figure}[!tbp] 
	\centering
	\small
	\includegraphics[width=0.99 \linewidth]{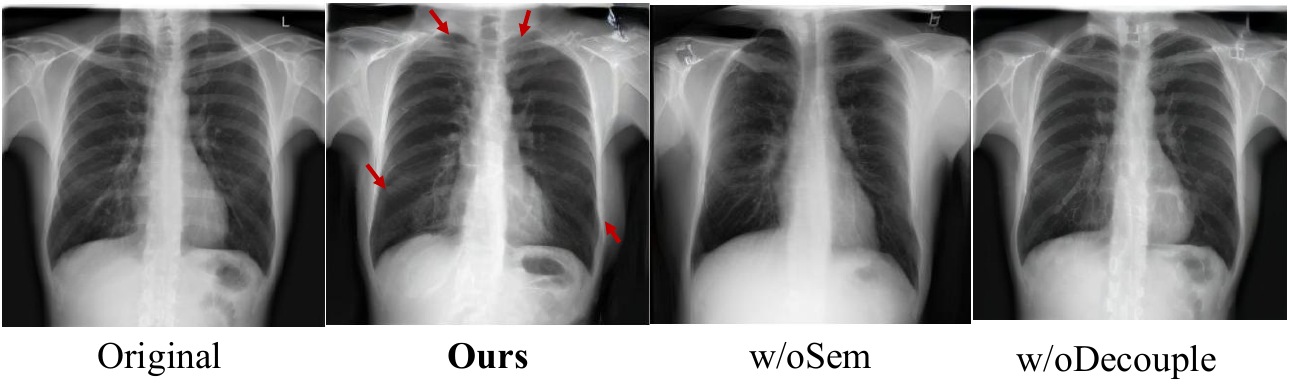}
	\vspace{-2mm}
	\caption{
		Qualitative comparison of different variant models.  
		The models are described in Figure~\ref{fig:ablation} caption.
		Red arrow denotes the modified identity-related features.
		Best to view by zooming-in.
	}
	\vspace{-6mm}
	\label{fig:visual_semantics}
\end{figure}

\subsection{Model Analysis}

\textbf{Framework-Level Ablation Study.}  
As shown in Figure~\ref{fig:ablation}~(\textit{Left}), by removing the semantic branch, the AUROC of the resulted model `w/oSem' dramatically drops by over 8\%, at CMC-R1=5\%.
On the other hand, without the identity-semantics decoupling mechanism, the resulting model `w/oDecouple' leads to about 15\% CMC-R1 increase, for achieving the similar AUROC performance, since substantial identity cues are leaked from the vanilla medical features of MFMs.
We further illustrate the protected images from different models.
As shown in Figure~\ref{fig:visual_semantics}, 	our results effectively modify identity-related features, such as the shape and location of the clavicle and chest contour.
The `w/oSem' model also removes these regions but significantly alters medical manifestations.  In contrast, the `w/oDecouple' model preserves medical features but fails to sufficiently suppress identity-related features, such as clavicle shape, due to residual identity information in the features from MFMs.
These results confirm that both medical semantics and identity-semantics decoupling are essential for our advanced medical DeID approach.

Besides, we attempt to remove ID-Blocking module, tuning privacy via activating more decoupled ID codes. This leads to degraded AUROC
(72.01\%,73.47\%,74.68\%), compared to using ID-Blocking (72.68\%,74.35\%,79.62\%) on MIMIC-X (10\%,20\%,40\% at
tack SR settings), due to the complete loss of pixel-wise details.

\textbf{Ablation Study on the Decoupling Strategy.}  
As shown in Figure~\ref{fig:ablation}~(\textit{Right}), omitting the codelength loss terms (`w/oCodeLengthLoss') fails to effectively remove identity information from MFM features, since the loose space cannot effectively decouple the identity and the semantics information. Moreover, removing the discrete code bottleneck (`w/oCodeSpace') further exacerbates identity leakage, leading to further inferior performance.

\begin{figure}[!tbp] 
	\centering
	\newcommand{\widthscalefive}{0.25}
	\tabcolsep = 1mm
	\scalebox{1}{
		\begin{tabular}{c}
			\includegraphics[width=0.98 \linewidth]{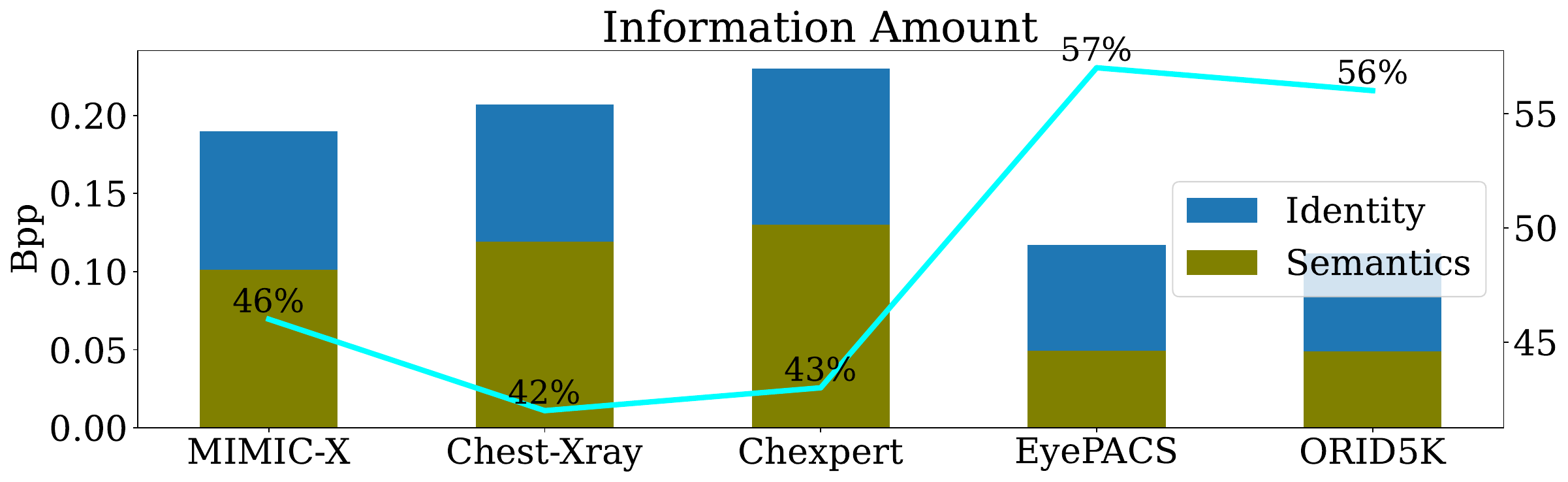} 
		\end{tabular}
	}
\vspace{-4mm}
	\caption{
		Comparison of semantic and identity information in terms of Bits-per-Pixel (bpp)~\cite{lin2006adaptive}, calculated as the feature codelength divided by the original image size.
	}
	\label{fig:quan_info}
	\vspace{-5mm}
\end{figure}

Furthermore, we quantitatively compare the overall and identity-related information in MFM features, as shown in Figure~\ref{fig:quan_info}. First, we notice that a significant portion is identity-related, i.e., around 44\% and 55\% for X-ray and fundus images.
Second, the average information amount of the X-ray dataset Chest-Xray is 0.23bpp, much higher than 0.11bpp achieved by the fundus dataset EyePACS.
This is aligned with the medical knowledge prior, that X-rays capture multiple organs and tissues, containing much complex information, than the fundus image that only focuses on eyes.
This proves that the learned codelength effectively describes the medical data characteristics.

\begin{figure}[!tbp] 
	\centering
	\small
	\newcommand{\widthscalefive}{0.25}
	\tabcolsep = 0mm
	\scalebox{1}{
		\begin{tabular}{cc}
			\includegraphics[width=0.48 \linewidth]{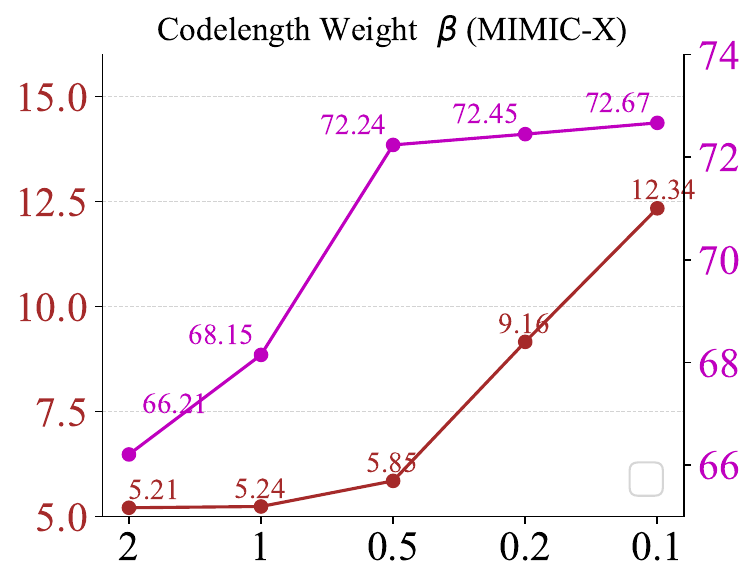}&
			\includegraphics[width=0.48 \linewidth]{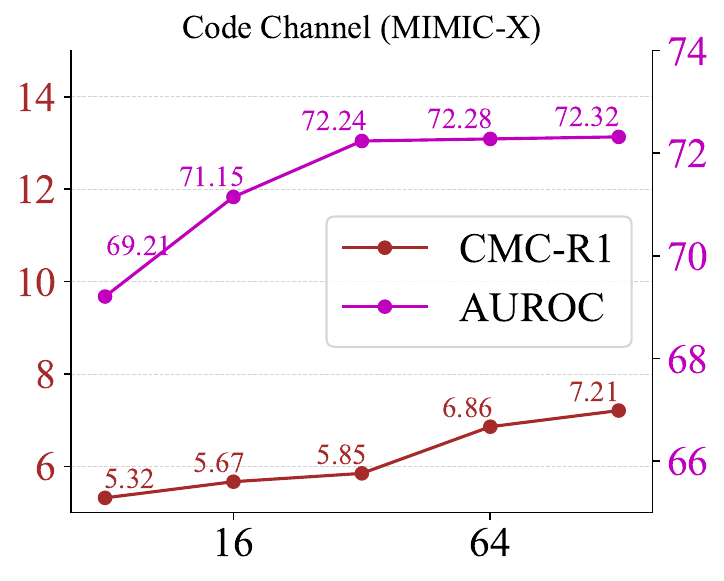}  \\
		\end{tabular}
	}
	\vspace{-2mm}
	\caption{
		(\textit{Left}) Impact of the rate-distortion weight $\beta$.
		(\textit{Right}:) Impact of the code dimension.
		All experiments are evaluated by masking 95\% identity-related regions, for a fair comparison.
	}
	\vspace{-4mm}
	\label{fig:ablation_hyper}
\end{figure}

Finally, we analyze the impact of the codelength loss weight $\beta$ and the latent code channel number. As shown in Figure~\ref{fig:ablation_hyper}~(\textit{Left}), reducing $\beta$ from 0.5 to 0.1 significantly increases CMC-R1 from 5.85\% to 12.34\%, as a loosely constrained code space fails to effectively decouple identity information. Conversely, increasing $\beta$ from 0.5 to 2 has little effect on CMC-R1 but reduces AUROC performance by approximately 6\%, as an overly strong constraint impairs semantic feature reconstruction. 
The number of code channels also influences performance, by tuning the information capacity of the latent code, as shown in Figure~\ref{fig:ablation_hyper}~(\textit{Right}). However, since $\beta$ directly regulates the code-length term, the impact of the channel number is limited.


\paragraph{Discussion with Label-Conditioned Diffusion Models.}  
These methods~\cite{huang2024chest,zhou2024ctrl,wang2024self} employ task-specific labels (e.g., disease labels or text reports) to synthesize images, which are limited to label-associated tasks. In contrast, our approach is task-agnostic and applicable to diverse tasks.
 Moreover, after fine-tuning our approach towards a single task, i.e., replacing the MFM with a supervised classification network, our method achieves 81.92\% AUROC at CMC-R1 = 0.30\%, surpassing the label-conditioned model, i.e., 80.79\% AUROC at CMC-R1 = 0.29\%. This confirms that our minimum-codelength representation also benefits the single-task setting, compared to the methods directly using the task labels guiding the diffusion procedure.

\textbf{Model Complexity.}  
All methods and our model comprise about 380M parameters, for a fair comparison.
Our inference time is 540 ms on an NVIDIA 4090 GPU, which is similar to MAE (526ms), but slower than Privacy-Net (120 ms), ID-Adv (122 ms), and Feat-Noise (124 ms), due to the multiple inference steps of diffusion procedure. Nonetheless, given the significant performance gains and that the medical imaging procedure itself is time-consuming, the running time is acceptable and does not hinder clinical workflows.
In the future, we will integrate the single-step diffusion technique~\cite{yin2024one} to accelerate the process.

\section{Conclusion, Future Works, and Other}
\textbf{Conclusion.}
We have presented DCM-DeID, a divide-and-conquer framework for medical image de-identification.
By leveraging pre-trained Medical Foundation Models and a minimum codelength-based feature decoupling strategy, our method effectively remove identity cues, while preserving medical task utility. Extensive evaluations demonstrate the superiority of our approach.
\textbf{Future Works.}
We will extend our approach to multi-slice images, such as those produced by Magnetic Resonance Imaging (MRI).
\textbf{Broader Impacts.}
Our DeID technique is designed for medical AI applications, aiding the human.
We emphasize that all rigorous clinical decisions must be made by human physicians using the original medical images. 
Furthermore, it is critical to enforce ethical guidelines, working in synergy with technological approaches to achieve medical privacy protection.

\noindent \textbf{Acknowledgment.} This work was supported by Shanghai Artificial Intelligence Laboratory, National Natural Science Foundation of China (Grant No.62225112), the Fundamental Research Funds for the Central Universities, National Key R\&D Program of China 2021YFE0206700, Shanghai Municipal Science and Technology Major Project (2021SHZDZX0102), and STCSM 22DZ2229005.

{	
    \small
    \bibliographystyle{ieeenat_fullname}
    \bibliography{main}

\begin{thebibliography}{126}
\providecommand{\natexlab}[1]{#1}
\providecommand{\url}[1]{\texttt{#1}}
\expandafter\ifx\csname urlstyle\endcsname\relax
  \providecommand{\doi}[1]{doi: #1}\else
  \providecommand{\doi}{doi: \begingroup \urlstyle{rm}\Url}\fi

\bibitem[Aryanto et~al.(2012)Aryanto, Broekema, Oudkerk, and van
  Ooijen]{aryanto2012implementation}
Kadek~YE Aryanto, Andr{\'e} Broekema, Matthijs Oudkerk, and Peter~MA van
  Ooijen.
\newblock Implementation of an anonymisation tool for clinical trials using a
  clinical trial processor integrated with an existing trial patient data
  information system.
\newblock \emph{European radiology}, 22:\penalty0 144--151, 2012.

\bibitem[Ball{\'e} et~al.(2016)Ball{\'e}, Laparra, and
  Simoncelli]{balle2016end}
Johannes Ball{\'e}, Valero Laparra, and Eero~P Simoncelli.
\newblock End-to-end optimized image compression.
\newblock \emph{arXiv preprint arXiv:1611.01704}, 2016.

\bibitem[Ball{\'e} et~al.(2018)Ball{\'e}, Minnen, Singh, Hwang, and
  Johnston]{balle2018variational}
Johannes Ball{\'e}, David Minnen, Saurabh Singh, Sung~Jin Hwang, and Nick
  Johnston.
\newblock Variational image compression with a scale hyperprior.
\newblock \emph{arXiv preprint arXiv:1802.01436}, 2018.

\bibitem[Barron et~al.(1998)Barron, Rissanen, and Yu]{barron1998minimum}
Andrew Barron, Jorma Rissanen, and Bin Yu.
\newblock The minimum description length principle in coding and modeling.
\newblock \emph{IEEE transactions on information theory}, 44\penalty0
  (6):\penalty0 2743--2760, 1998.

\bibitem[Berger(2003)]{berger2003rate}
Toby Berger.
\newblock Rate-distortion theory.
\newblock \emph{Wiley Encyclopedia of Telecommunications}, 2003.

\bibitem[Bernardes et~al.(2011)Bernardes, Serranho, and
  Lobo]{bernardes2011digital}
Rui Bernardes, Pedro Serranho, and Concei{\c{c}}{\~a}o Lobo.
\newblock Digital ocular fundus imaging: a review.
\newblock \emph{Ophthalmologica}, 226\penalty0 (4):\penalty0 161--181, 2011.

\bibitem[Bertels et~al.(2019)Bertels, Eelbode, Berman, Vandermeulen, Maes,
  Bisschops, and Blaschko]{bertels2019optimizing}
Jeroen Bertels, Tom Eelbode, Maxim Berman, Dirk Vandermeulen, Frederik Maes,
  Raf Bisschops, and Matthew~B Blaschko.
\newblock Optimizing the dice score and jaccard index for medical image
  segmentation: Theory and practice.
\newblock In \emph{Medical Image Computing and Computer Assisted
  Intervention--MICCAI 2019: 22nd International Conference, Shenzhen, China,
  October 13--17, 2019, Proceedings, Part II 22}, pages 92--100. Springer,
  2019.

\bibitem[Bhati et~al.(2023)Bhati, Gour, Khanna, and
  Ojha]{bhati2023discriminative}
Amit Bhati, Neha Gour, Pritee Khanna, and Aparajita Ojha.
\newblock Discriminative kernel convolution network for multi-label ophthalmic
  disease detection on imbalanced fundus image dataset.
\newblock \emph{Computers in Biology and Medicine}, 153:\penalty0 106519, 2023.

\bibitem[Bindschaedler et~al.(2017)Bindschaedler, Grubbs, Cash, Ristenpart, and
  Shmatikov]{bindschaedler2017tao}
Vincent Bindschaedler, Paul Grubbs, David Cash, Thomas Ristenpart, and Vitaly
  Shmatikov.
\newblock The tao of inference in privacy-protected databases.
\newblock \emph{Cryptology ePrint Archive}, 2017.

\bibitem[Bolle et~al.(2005)Bolle, Connell, Pankanti, Ratha, and
  Senior]{bolle2005relation}
Ruud~M Bolle, Jonathan~H Connell, Sharath Pankanti, Nalini~K Ratha, and
  Andrew~W Senior.
\newblock The relation between the roc curve and the cmc.
\newblock In \emph{Fourth IEEE workshop on automatic identification advanced
  technologies (AutoID'05)}, pages 15--20. IEEE, 2005.

\bibitem[Bradley(1997)]{bradley1997use}
Andrew~P Bradley.
\newblock The use of the area under the roc curve in the evaluation of machine
  learning algorithms.
\newblock \emph{Pattern recognition}, 30\penalty0 (7):\penalty0 1145--1159,
  1997.

\bibitem[Cao et~al.(2021)Cao, Liu, Wen, Xie, and Song]{cao2021personalized}
Jingyi Cao, Bo Liu, Yunqian Wen, Rong Xie, and Li Song.
\newblock Personalized and invertible face de-identification by disentangled
  identity information manipulation.
\newblock In \emph{Proceedings of the IEEE/CVF international conference on
  computer vision}, pages 3334--3342, 2021.

\bibitem[Chen et~al.(2021)Chen, Fan, and Panda]{chen2021crossvit}
Chun-Fu~Richard Chen, Quanfu Fan, and Rameswar Panda.
\newblock Crossvit: Cross-attention multi-scale vision transformer for image
  classification.
\newblock In \emph{Proceedings of the IEEE/CVF international conference on
  computer vision}, pages 357--366, 2021.

\bibitem[Chen et~al.(2016)Chen, Duan, Houthooft, Schulman, Sutskever, and
  Abbeel]{chen2016infogan}
Xi Chen, Yan Duan, Rein Houthooft, John Schulman, Ilya Sutskever, and Pieter
  Abbeel.
\newblock Infogan: Interpretable representation learning by information
  maximizing generative adversarial nets.
\newblock \emph{Advances in neural information processing systems}, 29, 2016.

\bibitem[Chen and Esmaeilzadeh(2024)]{chen2024generative}
Yan Chen and Pouyan Esmaeilzadeh.
\newblock Generative ai in medical practice: in-depth exploration of privacy
  and security challenges.
\newblock \emph{Journal of Medical Internet Research}, 26:\penalty0 e53008,
  2024.

\bibitem[Chen et~al.(2024)Chen, Sun, Tian, Jia, Zhang, Jiarui, Huang, Min,
  Zhai, and Zhang]{chen2024gaia}
Zijian Chen, Wei Sun, Yuan Tian, Jun Jia, Zicheng Zhang, Wang Jiarui, Ru Huang,
  Xiongkuo Min, Guangtao Zhai, and Wenjun Zhang.
\newblock Gaia: Rethinking action quality assessment for ai-generated videos.
\newblock \emph{Advances in Neural Information Processing Systems},
  37:\penalty0 40111--40144, 2024.

\bibitem[Cheng et~al.(2020)Cheng, Sun, Takeuchi, and Katto]{cheng2020learned}
Zhengxue Cheng, Heming Sun, Masaru Takeuchi, and Jiro Katto.
\newblock Learned image compression with discretized gaussian mixture
  likelihoods and attention modules.
\newblock In \emph{Proceedings of the IEEE/CVF conference on computer vision
  and pattern recognition}, pages 7939--7948, 2020.

\bibitem[Cohen and Mello(2019)]{cohen2019big}
I~Glenn Cohen and Michelle~M Mello.
\newblock Big data, big tech, and protecting patient privacy.
\newblock \emph{Jama}, 322\penalty0 (12):\penalty0 1141--1142, 2019.

\bibitem[Cover(1999)]{cover1999elements}
Thomas~M Cover.
\newblock \emph{Elements of information theory}.
\newblock John Wiley \& Sons, 1999.

\bibitem[Croft et~al.(2021)Croft, Sack, and Shi]{croft2021obfuscation}
William~L Croft, J{\"o}rg-R{\"u}diger Sack, and Wei Shi.
\newblock Obfuscation of images via differential privacy: From facial images to
  general images.
\newblock \emph{Peer-to-Peer Networking and Applications}, 14:\penalty0
  1705--1733, 2021.

\bibitem[Dai et~al.(2015)Dai, Saghafi, Wu, Konrad, and Ishwar]{dai2015towards}
Ji Dai, Behrouz Saghafi, Jonathan Wu, Janusz Konrad, and Prakash Ishwar.
\newblock Towards privacy-preserving recognition of human activities.
\newblock In \emph{2015 IEEE international conference on image processing
  (ICIP)}, pages 4238--4242. IEEE, 2015.

\bibitem[Deng et~al.(2009)Deng, Dong, Socher, Li, Li, and
  Fei-Fei]{deng2009imagenet}
Jia Deng, Wei Dong, Richard Socher, Li-Jia Li, Kai Li, and Li Fei-Fei.
\newblock Imagenet: A large-scale hierarchical image database.
\newblock In \emph{2009 IEEE conference on computer vision and pattern
  recognition}, pages 248--255. Ieee, 2009.

\bibitem[Deng et~al.(2019)Deng, Guo, Xue, and Zafeiriou]{deng2019arcface}
Jiankang Deng, Jia Guo, Niannan Xue, and Stefanos Zafeiriou.
\newblock Arcface: Additive angular margin loss for deep face recognition.
\newblock In \emph{Proceedings of the IEEE/CVF conference on computer vision
  and pattern recognition}, pages 4690--4699, 2019.

\bibitem[Deng et~al.(2020)Deng, Yang, Chen, Wen, and
  Tong]{deng2020disentangled}
Yu Deng, Jiaolong Yang, Dong Chen, Fang Wen, and Xin Tong.
\newblock Disentangled and controllable face image generation via 3d
  imitative-contrastive learning.
\newblock In \emph{Proceedings of the IEEE/CVF conference on computer vision
  and pattern recognition}, pages 5154--5163, 2020.

\bibitem[Dosovitskiy et~al.(2021)Dosovitskiy, Beyer, Kolesnikov, Weissenborn,
  Zhai, Unterthiner, Dehghani, Minderer, Heigold, Gelly, Uszkoreit, and
  Houlsby]{dosovitskiy2021an}
Alexey Dosovitskiy, Lucas Beyer, Alexander Kolesnikov, Dirk Weissenborn,
  Xiaohua Zhai, Thomas Unterthiner, Mostafa Dehghani, Matthias Minderer, Georg
  Heigold, Sylvain Gelly, Jakob Uszkoreit, and Neil Houlsby.
\newblock An image is worth 16x16 words: Transformers for image recognition at
  scale.
\newblock In \emph{International Conference on Learning Representations}, 2021.

\bibitem[Dugas et~al.(2015)Dugas, Jared, Jorge, and
  Cukierski]{diabetic-retinopathy-detection}
Emma Dugas, Jared, Jorge, and Will Cukierski.
\newblock Diabetic retinopathy detection.
\newblock \url{https://kaggle.com/competitions/diabetic-retinopathy-detection},
  2015.
\newblock Kaggle.

\bibitem[DuMont~Sch{\"u}tte et~al.(2021)DuMont~Sch{\"u}tte, Hetzel, Gatidis,
  Hepp, Dietz, Bauer, and Schwab]{dumont2021overcoming}
August DuMont~Sch{\"u}tte, J{\"u}rgen Hetzel, Sergios Gatidis, Tobias Hepp,
  Benedikt Dietz, Stefan Bauer, and Patrick Schwab.
\newblock Overcoming barriers to data sharing with medical image generation: a
  comprehensive evaluation.
\newblock \emph{NPJ digital medicine}, 4\penalty0 (1):\penalty0 141, 2021.

\bibitem[Dwork(2006)]{dwork2006differential}
Cynthia Dwork.
\newblock Differential privacy.
\newblock In \emph{International colloquium on automata, languages, and
  programming}, pages 1--12. Springer, 2006.

\bibitem[Fan et~al.(2024)Fan, Yang, Qi, Zhang, Liao, Wen, Wang, Wang, Xia, Wu,
  et~al.]{fan2024deep}
Weijie Fan, Yi Yang, Jing Qi, Qichuan Zhang, Cuiwei Liao, Li Wen, Shuang Wang,
  Guangxian Wang, Yu Xia, Qihua Wu, et~al.
\newblock A deep-learning-based framework for identifying and localizing
  multiple abnormalities and assessing cardiomegaly in chest x-ray.
\newblock \emph{Nature Communications}, 15\penalty0 (1):\penalty0 1347, 2024.

\bibitem[Fang et~al.(2022)Fang, Li, Wu, Fu, Sun, Son, Yu, Zhang, Yuan, Bian,
  et~al.]{fang2022refuge2}
Huihui Fang, Fei Li, Junde Wu, Huazhu Fu, Xu Sun, Jaemin Son, Shuang Yu, Menglu
  Zhang, Chenglang Yuan, Cheng Bian, et~al.
\newblock Refuge2 challenge: A treasure trove for multi-dimension analysis and
  evaluation in glaucoma screening.
\newblock \emph{arXiv preprint arXiv:2202.08994}, 2022.

\bibitem[Fischl(2012)]{fischl2012freesurfer}
Bruce Fischl.
\newblock Freesurfer.
\newblock \emph{Neuroimage}, 62\penalty0 (2):\penalty0 774--781, 2012.

\bibitem[Gaudio et~al.(2023)Gaudio, Smailagic, Faloutsos, Mohan, Johnson, Liu,
  Costa, and Campilho]{gaudio2023deepfixcx}
Alex Gaudio, Asim Smailagic, Christos Faloutsos, Shreshta Mohan, Elvin Johnson,
  Yuhao Liu, Pedro Costa, and Aur{\'e}lio Campilho.
\newblock Deepfixcx: Explainable privacy-preserving image compression for
  medical image analysis.
\newblock \emph{Wiley interdisciplinary reviews: Data mining and knowledge
  discovery}, 13\penalty0 (4):\penalty0 e1495, 2023.

\bibitem[Ghosh et~al.(2012)Ghosh, Gorgolewski, et~al.]{pydeface}
Satrajit Ghosh, Chris Gorgolewski, et~al.
\newblock pydeface: A tool to remove facial features from mri images, 2012.

\bibitem[Gr{\"u}nwald(2007)]{grunwald2007minimum}
Peter~D Gr{\"u}nwald.
\newblock \emph{The minimum description length principle}.
\newblock MIT press, 2007.

\bibitem[Gu et~al.(2020)Gu, Luo, Ryoo, and Lee]{gu2020password}
Xiuye Gu, Weixin Luo, Michael~S Ryoo, and Yong~Jae Lee.
\newblock Password-conditioned anonymization and deanonymization with face
  identity transformers.
\newblock In \emph{European conference on computer vision}, pages 727--743.
  Springer, 2020.

\bibitem[Gueziec and Kazanzides(1998)]{1998Anatomy}
A Gueziec and P Kazanzides.
\newblock Anatomy-based registration of ct-scan and intraoperative x-ray images
  for guiding a surgical robot.
\newblock \emph{IEEE Transactions on Medical Imaging}, 17\penalty0
  (5):\penalty0 715--728, 1998.

\bibitem[He et~al.(2016)He, Zhang, Ren, and Sun]{he2016deep}
Kaiming He, Xiangyu Zhang, Shaoqing Ren, and Jian Sun.
\newblock Deep residual learning for image recognition.
\newblock In \emph{Proceedings of the IEEE conference on computer vision and
  pattern recognition}, pages 770--778, 2016.

\bibitem[He et~al.(2022)He, Chen, Xie, Li, Doll{\'a}r, and
  Girshick]{he2022masked}
Kaiming He, Xinlei Chen, Saining Xie, Yanghao Li, Piotr Doll{\'a}r, and Ross
  Girshick.
\newblock Masked autoencoders are scalable vision learners.
\newblock In \emph{Proceedings of the IEEE/CVF conference on computer vision
  and pattern recognition}, pages 16000--16009, 2022.

\bibitem[He et~al.(2021)He, Luo, Wang, Wang, Li, and Jiang]{he2021transreid}
Shuting He, Hao Luo, Pichao Wang, Fan Wang, Hao Li, and Wei Jiang.
\newblock Transreid: Transformer-based object re-identification.
\newblock In \emph{Proceedings of the IEEE/CVF international conference on
  computer vision}, pages 15013--15022, 2021.

\bibitem[Hermans et~al.(2017)Hermans, Beyer, and Leibe]{hermans2017defense}
Alexander Hermans, Lucas Beyer, and Bastian Leibe.
\newblock In defense of the triplet loss for person re-identification.
\newblock \emph{arXiv preprint arXiv:1703.07737}, 2017.

\bibitem[Higgins et~al.(2017)Higgins, Matthey, Pal, Burgess, Glorot, Botvinick,
  Mohamed, and Lerchner]{higgins2017beta}
Irina Higgins, Loic Matthey, Arka Pal, Christopher Burgess, Xavier Glorot,
  Matthew Botvinick, Shakir Mohamed, and Alexander Lerchner.
\newblock beta-vae: Learning basic visual concepts with a constrained
  variational framework.
\newblock In \emph{International conference on learning representations}, 2017.

\bibitem[Hill et~al.(2016)Hill, Zhou, Saul, and Shacham]{hill2016effectiveness}
Steven Hill, Zhimin Zhou, Lawrence Saul, and Hovav Shacham.
\newblock On the (in) effectiveness of mosaicing and blurring as tools for
  document redaction.
\newblock \emph{Proceedings on Privacy Enhancing Technologies}, 2016.

\bibitem[Ho et~al.(2020)Ho, Jain, and Abbeel]{ho2020denoising}
Jonathan Ho, Ajay Jain, and Pieter Abbeel.
\newblock Denoising diffusion probabilistic models.
\newblock \emph{Advances in neural information processing systems},
  33:\penalty0 6840--6851, 2020.

\bibitem[Hong et~al.(2021)Hong, Marinescu, Dalca, Bonkhoff, Bretzner, Rost, and
  Golland]{hong20213d}
Sungmin Hong, Razvan Marinescu, Adrian~V Dalca, Anna~K Bonkhoff, Martin
  Bretzner, Natalia~S Rost, and Polina Golland.
\newblock 3d-stylegan: A style-based generative adversarial network for
  generative modeling of three-dimensional medical images.
\newblock In \emph{Deep Generative Models, and Data Augmentation, Labelling,
  and Imperfections: First Workshop, DGM4MICCAI 2021, and First Workshop, DALI
  2021, Held in Conjunction with MICCAI 2021, Strasbourg, France, October 1,
  2021, Proceedings 1}, pages 24--34. Springer, 2021.

\bibitem[Hoopes et~al.(2022)Hoopes, Mora, Dalca, Fischl, and
  Hoffmann]{hoopes2022synthstrip}
Andrew Hoopes, Jocelyn~S Mora, Adrian~V Dalca, Bruce Fischl, and Malte
  Hoffmann.
\newblock Synthstrip: skull-stripping for any brain image.
\newblock \emph{NeuroImage}, 260:\penalty0 119474, 2022.

\bibitem[Hu et~al.(2025{\natexlab{a}})Hu, He, Zhong, Lu, Zhang, Zhai, and
  Wang]{hu2025varfvv}
Qiang Hu, Qihan He, Houqiang Zhong, Guo Lu, Xiaoyun Zhang, Guangtao Zhai, and
  Yanfeng Wang.
\newblock Varfvv: View-adaptive real-time interactive free-view video streaming
  with edge computing.
\newblock \emph{arXiv preprint arXiv:2501.13630}, 2025{\natexlab{a}}.

\bibitem[Hu et~al.(2025{\natexlab{b}})Hu, Zheng, Zhong, Fu, Song, Zhang, Zhai,
  and Wang]{hu20254dgc}
Qiang Hu, Zihan Zheng, Houqiang Zhong, Sihua Fu, Li Song, Xiaoyun Zhang,
  Guangtao Zhai, and Yanfeng Wang.
\newblock 4dgc: Rate-aware 4d gaussian compression for efficient streamable
  free-viewpoint video.
\newblock In \emph{Proceedings of the Computer Vision and Pattern Recognition
  Conference}, pages 875--885, 2025{\natexlab{b}}.

\bibitem[Hu et~al.(2025{\natexlab{c}})Hu, Zhong, Zheng, Zhang, Cheng, Song,
  Zhai, and Wang]{hu2025vrvvc}
Qiang Hu, Houqiang Zhong, Zihan Zheng, Xiaoyun Zhang, Zhengxue Cheng, Li Song,
  Guangtao Zhai, and Yanfeng Wang.
\newblock Vrvvc: Variable-rate nerf-based volumetric video compression.
\newblock In \emph{Proceedings of the AAAI Conference on Artificial
  Intelligence}, pages 3563--3571, 2025{\natexlab{c}}.

\bibitem[Huang et~al.(2024)Huang, Gao, Huang, Jiao, Li, Wang, and
  Guo]{huang2024chest}
Peng Huang, Xue Gao, Lihong Huang, Jing Jiao, Xiaokang Li, Yuanyuan Wang, and
  Yi Guo.
\newblock Chest-diffusion: a light-weight text-to-image model for report-to-cxr
  generation.
\newblock In \emph{2024 IEEE International Symposium on Biomedical Imaging
  (ISBI)}, pages 1--5. IEEE, 2024.

\bibitem[Isensee et~al.(2021)Isensee, Jaeger, Kohl, Petersen, and
  Maier-Hein]{isensee2021nnu}
Fabian Isensee, Paul~F Jaeger, Simon~AA Kohl, Jens Petersen, and Klaus~H
  Maier-Hein.
\newblock nnu-net: a self-configuring method for deep learning-based biomedical
  image segmentation.
\newblock \emph{Nature methods}, 18\penalty0 (2):\penalty0 203--211, 2021.

\bibitem[Jang et~al.(2016)Jang, Gu, and Poole]{jang2016categorical}
Eric Jang, Shixiang Gu, and Ben Poole.
\newblock Categorical reparameterization with gumbel-softmax.
\newblock \emph{arXiv preprint arXiv:1611.01144}, 2016.

\bibitem[Johnson et~al.(2019)Johnson, Pollard, Berkowitz, Greenbaum, Lungren,
  Deng, Mark, and Horng]{johnson2019mimic}
Alistair~EW Johnson, Tom~J Pollard, Seth~J Berkowitz, Nathaniel~R Greenbaum,
  Matthew~P Lungren, Chih-ying Deng, Roger~G Mark, and Steven Horng.
\newblock Mimic-cxr, a de-identified publicly available database of chest
  radiographs with free-text reports.
\newblock \emph{Scientific data}, 6\penalty0 (1):\penalty0 317, 2019.

\bibitem[Kaissis et~al.(2020)Kaissis, Makowski, R{\"u}ckert, and
  Braren]{kaissis2020secure}
Georgios~A Kaissis, Marcus~R Makowski, Daniel R{\"u}ckert, and Rickmer~F
  Braren.
\newblock Secure, privacy-preserving and federated machine learning in medical
  imaging.
\newblock \emph{Nature Machine Intelligence}, 2\penalty0 (6):\penalty0
  305--311, 2020.

\bibitem[Karras et~al.(2019)Karras, Laine, and Aila]{karras2019style}
Tero Karras, Samuli Laine, and Timo Aila.
\newblock A style-based generator architecture for generative adversarial
  networks.
\newblock In \emph{Proceedings of the IEEE/CVF conference on computer vision
  and pattern recognition}, pages 4401--4410, 2019.

\bibitem[Karras et~al.(2020)Karras, Laine, Aittala, Hellsten, Lehtinen, and
  Aila]{karras2020analyzing}
Tero Karras, Samuli Laine, Miika Aittala, Janne Hellsten, Jaakko Lehtinen, and
  Timo Aila.
\newblock Analyzing and improving the image quality of stylegan.
\newblock In \emph{Proceedings of the IEEE/CVF conference on computer vision
  and pattern recognition}, pages 8110--8119, 2020.

\bibitem[Kim et~al.(2021)Kim, Dolz, Jodoin, and Desrosiers]{kim2021privacy}
Bach~Ngoc Kim, Jose Dolz, Pierre-Marc Jodoin, and Christian Desrosiers.
\newblock Privacy-net: an adversarial approach for identity-obfuscated
  segmentation of medical images.
\newblock \emph{IEEE Transactions on Medical Imaging}, 40\penalty0
  (7):\penalty0 1737--1749, 2021.

\bibitem[Koch and Larrabee(2013)]{koch2013patient}
Cody~A Koch and Wayne~F Larrabee.
\newblock Patient privacy, photographs, and publication.
\newblock \emph{JAMA facial plastic surgery}, 15\penalty0 (5):\penalty0
  335--336, 2013.

\bibitem[Kumar et~al.(2021)Kumar, Singh, Lakshmanan, Saxena, and
  Shrivastava]{kumar2021novel}
Abhinav Kumar, Sanjay~Kumar Singh, K Lakshmanan, Sonal Saxena, and Sameer
  Shrivastava.
\newblock A novel cloud-assisted secure deep feature classification framework
  for cancer histopathology images.
\newblock \emph{ACM Transactions on Internet Technology (TOIT)}, 21\penalty0
  (2):\penalty0 1--22, 2021.

\bibitem[Lee et~al.(2025)Lee, Youn, Kim, Kim, and Yoon]{lee2025cxr}
Seowoo Lee, Jiwon Youn, Hyungjin Kim, Mansu Kim, and Soon~Ho Yoon.
\newblock Cxr-llava: a multimodal large language model for interpreting chest
  x-ray images.
\newblock \emph{European Radiology}, pages 1--13, 2025.

\bibitem[Li et~al.(2023)Li, Zhang, Wu, Sun, Min, Liu, Zhai, and
  Lin]{li2023agiqa}
Chunyi Li, Zicheng Zhang, Haoning Wu, Wei Sun, Xiongkuo Min, Xiaohong Liu,
  Guangtao Zhai, and Weisi Lin.
\newblock Agiqa-3k: An open database for ai-generated image quality assessment.
\newblock \emph{IEEE Transactions on Circuits and Systems for Video
  Technology}, 34\penalty0 (8):\penalty0 6833--6846, 2023.

\bibitem[Li et~al.(2025{\natexlab{a}})Li, Li, Zhang, Tian, Jia, Liu, Min, Wang,
  Duan, Chen, et~al.]{li2025information}
Chunyi Li, Xiaozhe Li, Zicheng Zhang, Yuan Tian, Ziheng Jia, Xiaohong Liu,
  Xiongkuo Min, Jia Wang, Haodong Duan, Kai Chen, et~al.
\newblock Information density principle for mllm benchmarks.
\newblock \emph{arXiv preprint arXiv:2503.10079}, 2025{\natexlab{a}}.

\bibitem[Li et~al.(2025{\natexlab{b}})Li, Tian, Ling, Zhang, Duan, Wu, Jia,
  Liu, Min, Lu, et~al.]{li2025image}
Chunyi Li, Yuan Tian, Xiaoyue Ling, Zicheng Zhang, Haodong Duan, Haoning Wu,
  Ziheng Jia, Xiaohong Liu, Xiongkuo Min, Guo Lu, et~al.
\newblock Image quality assessment: From human to machine preference.
\newblock In \emph{Proceedings of the Computer Vision and Pattern Recognition
  Conference}, pages 7570--7581, 2025{\natexlab{b}}.

\bibitem[Li et~al.(2025{\natexlab{c}})Li, Xiao, Zhang, Wen, Zhang, Tian, Zhu,
  Liu, Cheng, Lin, et~al.]{li2025perceptual}
Chunyi Li, Jiaohao Xiao, Jianbo Zhang, Farong Wen, Zicheng Zhang, Yuan Tian,
  Xiangyang Zhu, Xiaohong Liu, Zhengxue Cheng, Weisi Lin, et~al.
\newblock Perceptual quality assessment for embodied ai.
\newblock \emph{arXiv preprint arXiv:2505.16815}, 2025{\natexlab{c}}.

\bibitem[Li et~al.(2020)Li, Singh, Ojha, and Lee]{li2020mixnmatch}
Yuheng Li, Krishna~Kumar Singh, Utkarsh Ojha, and Yong~Jae Lee.
\newblock Mixnmatch: Multifactor disentanglement and encoding for conditional
  image generation.
\newblock In \emph{Proceedings of the IEEE/CVF conference on computer vision
  and pattern recognition}, pages 8039--8048, 2020.

\bibitem[Lian et~al.(2021)Lian, Liu, Zhang, Gao, Liu, Zhang, and
  Yu]{lian2021structure}
Jie Lian, Jingyu Liu, Shu Zhang, Kai Gao, Xiaoqing Liu, Dingwen Zhang, and
  Yizhou Yu.
\newblock A structure-aware relation network for thoracic diseases detection
  and segmentation.
\newblock \emph{IEEE Transactions on Medical Imaging}, 40\penalty0
  (8):\penalty0 2042--2052, 2021.

\bibitem[Lin and Dong(2006)]{lin2006adaptive}
Weisi Lin and Li Dong.
\newblock Adaptive downsampling to improve image compression at low bit rates.
\newblock \emph{IEEE Transactions on Image Processing}, 15\penalty0
  (9):\penalty0 2513--2521, 2006.

\bibitem[Loshchilov(2017)]{loshchilov2017decoupled}
I Loshchilov.
\newblock Decoupled weight decay regularization.
\newblock \emph{arXiv preprint arXiv:1711.05101}, 2017.

\bibitem[Loshchilov and Hutter(2016)]{loshchilov2016sgdr}
Ilya Loshchilov and Frank Hutter.
\newblock Sgdr: Stochastic gradient descent with warm restarts.
\newblock \emph{arXiv preprint arXiv:1608.03983}, 2016.

\bibitem[Marsh(2013)]{marsh2013introduction}
Charles Marsh.
\newblock Introduction to continuous entropy.
\newblock \emph{Department of Computer Science, Princeton University}, 1034,
  2013.

\bibitem[Maximov et~al.(2020)Maximov, Elezi, and
  Leal-Taix{\'e}]{maximov2020ciagan}
Maxim Maximov, Ismail Elezi, and Laura Leal-Taix{\'e}.
\newblock Ciagan: Conditional identity anonymization generative adversarial
  networks.
\newblock In \emph{Proceedings of the IEEE/CVF conference on computer vision
  and pattern recognition}, pages 5447--5456, 2020.

\bibitem[Mentzer et~al.(2018)Mentzer, Agustsson, Tschannen, Timofte, and
  Van~Gool]{mentzer2018conditional}
Fabian Mentzer, Eirikur Agustsson, Michael Tschannen, Radu Timofte, and Luc
  Van~Gool.
\newblock Conditional probability models for deep image compression.
\newblock In \emph{Proceedings of the IEEE conference on computer vision and
  pattern recognition}, pages 4394--4402, 2018.

\bibitem[Mentzer et~al.(2023)Mentzer, Minnen, Agustsson, and
  Tschannen]{mentzer2023finite}
Fabian Mentzer, David Minnen, Eirikur Agustsson, and Michael Tschannen.
\newblock Finite scalar quantization: Vq-vae made simple.
\newblock \emph{arXiv preprint arXiv:2309.15505}, 2023.

\bibitem[Minnen et~al.(2018)Minnen, Ball{\'e}, and Toderici]{minnen2018joint}
David Minnen, Johannes Ball{\'e}, and George~D Toderici.
\newblock Joint autoregressive and hierarchical priors for learned image
  compression.
\newblock \emph{Advances in neural information processing systems}, 31, 2018.

\bibitem[Monteiro et~al.(2017)Monteiro, Costa, and Oliveira]{2017A}
Eriksson Monteiro, Carlos Costa, and José~Luís Oliveira.
\newblock A de-identification pipeline for ultrasound medical images in dicom
  format.
\newblock \emph{Journal of Medical Systems}, 41\penalty0 (5):\penalty0 89,
  2017.

\bibitem[Moor et~al.(2023)Moor, Banerjee, Abad, Krumholz, Leskovec, Topol, and
  Rajpurkar]{moor2023foundation}
Michael Moor, Oishi Banerjee, Zahra Shakeri~Hossein Abad, Harlan~M Krumholz,
  Jure Leskovec, Eric~J Topol, and Pranav Rajpurkar.
\newblock Foundation models for generalist medical artificial intelligence.
\newblock \emph{Nature}, 616\penalty0 (7956):\penalty0 259--265, 2023.

\bibitem[Ni et~al.(2024)Ni, Chen, Zhai, Tang, and Wang]{ni2024context}
Zhenliang Ni, Xinghao Chen, Yingjie Zhai, Yehui Tang, and Yunhe Wang.
\newblock Context-guided spatial feature reconstruction for efficient semantic
  segmentation.
\newblock In \emph{European Conference on Computer Vision}, pages 239--255.
  Springer, 2024.

\bibitem[Packh{\"a}user et~al.(2022)Packh{\"a}user, G{\"u}ndel, M{\"u}nster,
  Syben, Christlein, and Maier]{packhauser2022deep}
Kai Packh{\"a}user, Sebastian G{\"u}ndel, Nicolas M{\"u}nster, Christopher
  Syben, Vincent Christlein, and Andreas Maier.
\newblock Deep learning-based patient re-identification is able to exploit the
  biometric nature of medical chest x-ray data.
\newblock \emph{Scientific Reports}, 12\penalty0 (1):\penalty0 14851, 2022.

\bibitem[Packh{\"a}user et~al.(2023)Packh{\"a}user, G{\"u}ndel, Thamm,
  Denzinger, and Maier]{packhauser2023deep}
Kai Packh{\"a}user, Sebastian G{\"u}ndel, Florian Thamm, Felix Denzinger, and
  Andreas Maier.
\newblock Deep learning-based anonymization of chest radiographs: a
  utility-preserving measure for patient privacy.
\newblock In \emph{International Conference on Medical Image Computing and
  Computer-Assisted Intervention}, pages 262--272. Springer, 2023.

\bibitem[Paillier(1999)]{paillier1999public}
Pascal Paillier.
\newblock Public-key cryptosystems based on composite degree residuosity
  classes.
\newblock In \emph{International conference on the theory and applications of
  cryptographic techniques}, pages 223--238. Springer, 1999.

\bibitem[Papineni et~al.(2002)Papineni, Roukos, Ward, and
  Zhu]{papineni2002bleu}
Kishore Papineni, Salim Roukos, Todd Ward, and Wei-Jing Zhu.
\newblock Bleu: a method for automatic evaluation of machine translation.
\newblock In \emph{Proceedings of the 40th annual meeting of the Association
  for Computational Linguistics}, pages 311--318, 2002.

\bibitem[Paszke et~al.(2019)Paszke, Gross, Massa, Lerer, Bradbury, Chanan,
  Killeen, Lin, Gimelshein, Antiga, et~al.]{paszke2019pytorch}
Adam Paszke, Sam Gross, Francisco Massa, Adam Lerer, James Bradbury, Gregory
  Chanan, Trevor Killeen, Zeming Lin, Natalia Gimelshein, Luca Antiga, et~al.
\newblock Pytorch: An imperative style, high-performance deep learning library.
\newblock \emph{Advances in neural information processing systems}, 32, 2019.

\bibitem[Price and Cohen(2019)]{2019Privacy}
W.~Nicholson Price and I.~Glenn Cohen.
\newblock Privacy in the age of medical big data.
\newblock \emph{Nature Medicine}, 25\penalty0 (1):\penalty0 37--43, 2019.

\bibitem[Ra et~al.(2013)Ra, Govindan, and Ortega]{ra2013p3}
Moo-Ryong Ra, Ramesh Govindan, and Antonio Ortega.
\newblock P3: Toward $\{$Privacy-Preserving$\}$ photo sharing.
\newblock In \emph{10th USENIX Symposium on Networked Systems Design and
  Implementation (NSDI 13)}, pages 515--528, 2013.

\bibitem[Radford et~al.(2021)Radford, Kim, Hallacy, Ramesh, Goh, Agarwal,
  Sastry, Askell, Mishkin, Clark, et~al.]{radford2021learning}
Alec Radford, Jong~Wook Kim, Chris Hallacy, Aditya Ramesh, Gabriel Goh,
  Sandhini Agarwal, Girish Sastry, Amanda Askell, Pamela Mishkin, Jack Clark,
  et~al.
\newblock Learning transferable visual models from natural language
  supervision.
\newblock In \emph{International conference on machine learning}, pages
  8748--8763. PMLR, 2021.

\bibitem[Reynolds et~al.(2009)]{reynolds2009gaussian}
Douglas~A Reynolds et~al.
\newblock Gaussian mixture models.
\newblock \emph{Encyclopedia of biometrics}, 741\penalty0 (659-663):\penalty0
  3, 2009.

\bibitem[Rodr{\'\i}guez~Gonz{\'a}lez et~al.(2010)Rodr{\'\i}guez~Gonz{\'a}lez,
  Carpenter, van Hemert, and Wardlaw]{rodriguez2010open}
David Rodr{\'\i}guez~Gonz{\'a}lez, Trevor Carpenter, Jano~I van Hemert, and
  Joanna Wardlaw.
\newblock An open source toolkit for medical imaging de-identification.
\newblock \emph{European radiology}, 20:\penalty0 1896--1904, 2010.

\bibitem[Rombach et~al.(2022)Rombach, Blattmann, Lorenz, Esser, and
  Ommer]{rombach2022high}
Robin Rombach, Andreas Blattmann, Dominik Lorenz, Patrick Esser, and Bj{\"o}rn
  Ommer.
\newblock High-resolution image synthesis with latent diffusion models.
\newblock In \emph{Proceedings of the IEEE/CVF conference on computer vision
  and pattern recognition}, pages 10684--10695, 2022.

\bibitem[Seyyed-Kalantari et~al.(2021)Seyyed-Kalantari, Zhang, McDermott, Chen,
  and Ghassemi]{seyyed2021underdiagnosis}
Laleh Seyyed-Kalantari, Haoran Zhang, Matthew~BA McDermott, Irene~Y Chen, and
  Marzyeh Ghassemi.
\newblock Underdiagnosis bias of artificial intelligence algorithms applied to
  chest radiographs in under-served patient populations.
\newblock \emph{Nature medicine}, 27\penalty0 (12):\penalty0 2176--2182, 2021.

\bibitem[Shao et~al.(2022)Shao, Yang, Lin, Lin, Chen, Yang, and
  Zhao]{shao2022rethinking}
Huajie Shao, Yifei Yang, Haohong Lin, Longzhong Lin, Yizhuo Chen, Qinmin Yang,
  and Han Zhao.
\newblock Rethinking controllable variational autoencoders.
\newblock In \emph{Proceedings of the IEEE/CVF Conference on Computer Vision
  and Pattern Recognition}, pages 19250--19259, 2022.

\bibitem[Taitsman et~al.(2013)Taitsman, Grimm, and Agrawal]{2013Protecting}
Julie~K. Taitsman, Christi~Macrina Grimm, and Shantanu Agrawal.
\newblock Protecting patient privacy and data security.
\newblock \emph{New England Journal of Medicine}, 368\penalty0 (11):\penalty0
  977--979, 2013.

\bibitem[Tanner(2017)]{tanner2017our}
Adam Tanner.
\newblock \emph{Our bodies, our data: how companies make billions selling our
  medical records}.
\newblock Beacon Press, 2017.

\bibitem[Tian et~al.(2022{\natexlab{a}})Tian, Zhu, and Zhou]{tian2022fairness}
Huan Tian, Tianqing Zhu, and Wanlei Zhou.
\newblock Fairness and privacy preservation for facial images: Gan-based
  methods.
\newblock \emph{Computers \& Security}, 122:\penalty0 102902,
  2022{\natexlab{a}}.

\bibitem[Tian et~al.(2021)Tian, Lu, Min, Che, Zhai, Guo, and Gao]{tian2021self}
Yuan Tian, Guo Lu, Xiongkuo Min, Zhaohui Che, Guangtao Zhai, Guodong Guo, and
  Zhiyong Gao.
\newblock Self-conditioned probabilistic learning of video rescaling.
\newblock In \emph{Proceedings of the IEEE/CVF international conference on
  computer vision}, pages 4490--4499, 2021.

\bibitem[Tian et~al.(2022{\natexlab{b}})Tian, Yan, Zhai, Guo, and
  Gao]{tian2022ean}
Yuan Tian, Yichao Yan, Guangtao Zhai, Guodong Guo, and Zhiyong Gao.
\newblock Ean: event adaptive network for enhanced action recognition.
\newblock \emph{International Journal of Computer Vision}, 130\penalty0
  (10):\penalty0 2453--2471, 2022{\natexlab{b}}.

\bibitem[Tian et~al.(2023{\natexlab{a}})Tian, Lu, Zhai, and Gao]{tian2023non}
Yuan Tian, Guo Lu, Guangtao Zhai, and Zhiyong Gao.
\newblock Non-semantics suppressed mask learning for unsupervised video
  semantic compression.
\newblock In \emph{Proceedings of the IEEE/CVF International Conference on
  Computer Vision}, pages 13610--13622, 2023{\natexlab{a}}.

\bibitem[Tian et~al.(2023{\natexlab{b}})Tian, Yan, Zhai, Chen, and
  Gao]{tian2023clsa}
Yuan Tian, Yichao Yan, Guangtao Zhai, Li Chen, and Zhiyong Gao.
\newblock Clsa: a contrastive learning framework with selective aggregation for
  video rescaling.
\newblock \emph{IEEE Transactions on Image Processing}, 32:\penalty0
  1300--1314, 2023{\natexlab{b}}.

\bibitem[Tian et~al.(2024{\natexlab{a}})Tian, Lu, Yan, Zhai, Chen, and
  Gao]{tian2024coding}
Yuan Tian, Guo Lu, Yichao Yan, Guangtao Zhai, Li Chen, and Zhiyong Gao.
\newblock A coding framework and benchmark towards low-bitrate video
  understanding.
\newblock \emph{IEEE Transactions on Pattern Analysis and Machine
  Intelligence}, 2024{\natexlab{a}}.

\bibitem[Tian et~al.(2024{\natexlab{b}})Tian, Lu, and Zhai]{tian2024free}
Yuan Tian, Guo Lu, and Guangtao Zhai.
\newblock Free-vsc: Free semantics from visual foundation models for
  unsupervised video semantic compression.
\newblock In \emph{European Conference on Computer Vision}, pages 163--183.
  Springer, 2024{\natexlab{b}}.

\bibitem[Tian et~al.(2024{\natexlab{c}})Tian, Lu, and Zhai]{tian2024smc++}
Yuan Tian, Guo Lu, and Guangtao Zhai.
\newblock Smc++: Masked learning of unsupervised video semantic compression.
\newblock \emph{arXiv preprint arXiv:2406.04765}, 2024{\natexlab{c}}.

\bibitem[Tian et~al.(2025{\natexlab{a}})Tian, Ji, Zhang, Jiang, Li, Wang, and
  Zhai]{tian2025towards}
Yuan Tian, Kaiyuan Ji, Rongzhao Zhang, Yankai Jiang, Chunyi Li, Xiaosong Wang,
  and Guangtao Zhai.
\newblock Towards all-in-one medical image re-identification.
\newblock In \emph{Proceedings of the Computer Vision and Pattern Recognition
  Conference}, pages 30774--30786, 2025{\natexlab{a}}.

\bibitem[Tian et~al.(2025{\natexlab{b}})Tian, Wang, and Zhai]{tian2025medical}
Yuan Tian, Shuo Wang, and Guangtao Zhai.
\newblock Medical manifestation-aware de-identification.
\newblock In \emph{Proceedings of the AAAI Conference on Artificial
  Intelligence}, pages 26363--26372, 2025{\natexlab{b}}.

\bibitem[Tierney et~al.(2013)Tierney, Spiro, Bregler, and
  Subramanian]{tierney2013cryptagram}
Matt Tierney, Ian Spiro, Christoph Bregler, and Lakshminarayanan Subramanian.
\newblock Cryptagram: Photo privacy for online social media.
\newblock In \emph{Proceedings of the first ACM conference on Online social
  networks}, pages 75--88, 2013.

\bibitem[Tsui and Chan(2012)]{tsui2012automatic}
Gary Kin-wai Tsui and Tao Chan.
\newblock Automatic selective removal of embedded patient information from
  image content of dicom files.
\newblock \emph{American Journal of Roentgenology}, 198\penalty0 (4):\penalty0
  769--772, 2012.

\bibitem[Vishwamitra et~al.(2017)Vishwamitra, Knijnenburg, Hu, Kelly~Caine,
  et~al.]{vishwamitra2017blur}
Nishant Vishwamitra, Bart Knijnenburg, Hongxin Hu, Yifang~P Kelly~Caine, et~al.
\newblock Blur vs. block: Investigating the effectiveness of privacy-enhancing
  obfuscation for images.
\newblock In \emph{Proceedings of the IEEE Conference on Computer Vision and
  Pattern Recognition Workshops}, pages 39--47, 2017.

\bibitem[Wan et~al.(2024)Wan, Liu, Zhang, Fu, Wang, Cheng, Ma,
  Quilodr{\'a}n-Casas, and Arcucci]{wan2024med}
Zhongwei Wan, Che Liu, Mi Zhang, Jie Fu, Benyou Wang, Sibo Cheng, Lei Ma,
  C{\'e}sar Quilodr{\'a}n-Casas, and Rossella Arcucci.
\newblock Med-unic: Unifying cross-lingual medical vision-language pre-training
  by diminishing bias.
\newblock \emph{Advances in Neural Information Processing Systems}, 36, 2024.

\bibitem[Wang et~al.(2022{\natexlab{a}})Wang, Zhou, Wang, Vardhanabhuti, and
  Yu]{wang2022multi}
Fuying Wang, Yuyin Zhou, Shujun Wang, Varut Vardhanabhuti, and Lequan Yu.
\newblock Multi-granularity cross-modal alignment for generalized medical
  visual representation learning.
\newblock \emph{Advances in Neural Information Processing Systems},
  35:\penalty0 33536--33549, 2022{\natexlab{a}}.

\bibitem[Wang et~al.(2021)Wang, Liu, Shen, Wang, Li, Ye, Wu, Chen, Wang, Zhang,
  et~al.]{wang2021deep}
Guangyu Wang, Xiaohong Liu, Jun Shen, Chengdi Wang, Zhihuan Li, Linsen Ye,
  Xingwang Wu, Ting Chen, Kai Wang, Xuan Zhang, et~al.
\newblock A deep-learning pipeline for the diagnosis and discrimination of
  viral, non-viral and covid-19 pneumonia from chest x-ray images.
\newblock \emph{Nature biomedical engineering}, 5\penalty0 (6):\penalty0
  509--521, 2021.

\bibitem[Wang et~al.(2024{\natexlab{a}})Wang, Wang, Yu, Lu, Xiao, Sun, Liu,
  Zou, Gao, Yang, et~al.]{wang2024self}
Jinzhuo Wang, Kai Wang, Yunfang Yu, Yuxing Lu, Wenchao Xiao, Zhuo Sun, Fei Liu,
  Zixing Zou, Yuanxu Gao, Lei Yang, et~al.
\newblock Self-improving generative foundation model for synthetic medical
  image generation and clinical applications.
\newblock \emph{Nature Medicine}, pages 1--9, 2024{\natexlab{a}}.

\bibitem[Wang et~al.(2022{\natexlab{b}})Wang, Zhao, Peng, Zhu, Deng, Wang,
  Bilen, and You]{wang2022facemae}
Kai Wang, Bo Zhao, Xiangyu Peng, Zheng Zhu, Jiankang Deng, Xinchao Wang, Hakan
  Bilen, and Yang You.
\newblock Facemae: Privacy-preserving face recognition via masked autoencoders.
\newblock \emph{arXiv preprint arXiv:2205.11090}, 2022{\natexlab{b}}.

\bibitem[Wang et~al.(2024{\natexlab{b}})Wang, Zhao, Zhao, Wang, Zhang, Wang,
  Qiao, and Lyu]{wang2024semantic}
Shudong Wang, Zhiyuan Zhao, Yawu Zhao, Luqi Wang, Yuanyuan Zhang, Jiehuan Wang,
  Sibo Qiao, and Zhihan Lyu.
\newblock A semantic conditional diffusion model for enhanced personal privacy
  preservation in medical images.
\newblock \emph{IEEE Journal of Biomedical and Health Informatics},
  2024{\natexlab{b}}.

\bibitem[Wang et~al.(2017)Wang, Vong, Yang, and Wong]{wang2017encrypted}
Weiru Wang, Chi-Man Vong, Yilong Yang, and Pak-Kin Wong.
\newblock Encrypted image classification based on multilayer extreme learning
  machine.
\newblock \emph{Multidimensional Systems and Signal Processing}, 28\penalty0
  (3):\penalty0 851--865, 2017.

\bibitem[Wu et~al.(2024)Wu, Zhang, Zhang, Zhou, Zhou, and Fu]{wu2024mm}
Ruiqi Wu, Chenran Zhang, Jianle Zhang, Yi Zhou, Tao Zhou, and Huazhu Fu.
\newblock Mm-retinal: Knowledge-enhanced foundational pretraining with fundus
  image-text expertise.
\newblock In \emph{International Conference on Medical Image Computing and
  Computer-Assisted Intervention}, pages 722--732. Springer, 2024.

\bibitem[Xue et~al.(2021)Xue, Liu, Ding, Zhu, Ye, Song, and Zhou]{xue2021dp}
Hanyu Xue, Bo Liu, Ming Ding, Tianqing Zhu, Dayong Ye, Li Song, and Wanlei
  Zhou.
\newblock Dp-image: Differential privacy for image data in feature space.
\newblock \emph{arXiv preprint arXiv:2103.07073}, 2021.

\bibitem[Yin et~al.(2024)Yin, Gharbi, Zhang, Shechtman, Durand, Freeman, and
  Park]{yin2024one}
Tianwei Yin, Micha{\"e}l Gharbi, Richard Zhang, Eli Shechtman, Fredo Durand,
  William~T Freeman, and Taesung Park.
\newblock One-step diffusion with distribution matching distillation.
\newblock In \emph{Proceedings of the IEEE/CVF conference on computer vision
  and pattern recognition}, pages 6613--6623, 2024.

\bibitem[Yuan et~al.(2015{\natexlab{a}})Yuan, Korshunov, and
  Ebrahimi]{yuan2015privacy}
Lin Yuan, Pavel Korshunov, and Touradj Ebrahimi.
\newblock Privacy-preserving photo sharing based on a secure jpeg.
\newblock In \emph{2015 IEEE Conference on Computer Communications Workshops
  (INFOCOM WKSHPS)}, pages 185--190. IEEE, 2015{\natexlab{a}}.

\bibitem[Yuan et~al.(2015{\natexlab{b}})Yuan, Korshunov, and
  Ebrahimi]{yuan2015secure}
Lin Yuan, Pavel Korshunov, and Touradj Ebrahimi.
\newblock Secure jpeg scrambling enabling privacy in photo sharing.
\newblock In \emph{2015 11th IEEE International Conference and Workshops on
  Automatic Face and Gesture Recognition (FG)}, pages 1--6. IEEE,
  2015{\natexlab{b}}.

\bibitem[Zhai et~al.(2018)Zhai, Zhang, Chen, and He]{zhai2018autoencoder}
Junhai Zhai, Sufang Zhang, Junfen Chen, and Qiang He.
\newblock Autoencoder and its various variants.
\newblock In \emph{2018 IEEE international conference on systems, man, and
  cybernetics (SMC)}, pages 415--419. IEEE, 2018.

\bibitem[Zhang and Metaxas(2024)]{zhang2024challenges}
Shaoting Zhang and Dimitris Metaxas.
\newblock On the challenges and perspectives of foundation models for medical
  image analysis.
\newblock \emph{Medical image analysis}, 91:\penalty0 102996, 2024.

\bibitem[Zhang et~al.(2019)Zhang, Zhao, Qiao, Wang, and Li]{zhang2019adacos}
Xiao Zhang, Rui Zhao, Yu Qiao, Xiaogang Wang, and Hongsheng Li.
\newblock Adacos: Adaptively scaling cosine logits for effectively learning
  deep face representations.
\newblock In \emph{Proceedings of the IEEE/CVF Conference on Computer Vision
  and Pattern Recognition}, pages 10823--10832, 2019.

\bibitem[Zhang et~al.(2025)Zhang, Wang, Guo, Wen, Chen, Wang, Li, Sun, Zhou,
  Zhang, Yan, Jia, Xiao, Tian, Zhu, Zhang, Li, Liu, Min, Jia, and
  Zhai]{aibench}
Zicheng Zhang, Junying Wang, Yijin Guo, Farong Wen, Zijian Chen, Hanqing Wang,
  Wenzhe Li, Lu Sun, Yingjie Zhou, Jianbo Zhang, Bowen Yan, Ziheng Jia, Jiahao
  Xiao, Yuan Tian, Xiangyang Zhu, Kaiwei Zhang, Chunyi Li, Xiaohong Liu,
  Xiongkuo Min, Qi Jia, and Guangtao Zhai.
\newblock Aibench: Towards trustworthy evaluation under the 45° law.
\newblock \url{https://aiben.ch/}, 2025.

\bibitem[Zhou et~al.(2022)Zhou, Chen, Zhang, Luo, Wang, and
  Yu]{zhou2022generalized}
Hong-Yu Zhou, Xiaoyu Chen, Yinghao Zhang, Ruibang Luo, Liansheng Wang, and
  Yizhou Yu.
\newblock Generalized radiograph representation learning via cross-supervision
  between images and free-text radiology reports.
\newblock \emph{Nature Machine Intelligence}, 4\penalty0 (1):\penalty0 32--40,
  2022.

\bibitem[Zhou et~al.(2024)Zhou, Huang, Dou, Chen, Chang, Liu, Long, Zheng, Xu,
  Ren, et~al.]{zhou2024ctrl}
Xinrui Zhou, Yuhao Huang, Haoran Dou, Shijing Chen, Ao Chang, Jia Liu, Weiran
  Long, Jian Zheng, Erjiao Xu, Jie Ren, et~al.
\newblock Ctrl-genaug: Controllable generative augmentation for medical
  sequence classification.
\newblock \emph{arXiv preprint arXiv:2409.17091}, 2024.

\bibitem[Zhou et~al.(2023)Zhou, Chia, Wagner, Ayhan, Williamson, Struyven, Liu,
  Xu, Lozano, Woodward-Court, et~al.]{zhou2023foundation}
Yukun Zhou, Mark~A Chia, Siegfried~K Wagner, Murat~S Ayhan, Dominic~J
  Williamson, Robbert~R Struyven, Timing Liu, Moucheng Xu, Mateo~G Lozano,
  Peter Woodward-Court, et~al.
\newblock A foundation model for generalizable disease detection from retinal
  images.
\newblock \emph{Nature}, 622\penalty0 (7981):\penalty0 156--163, 2023.

\bibitem[Zhu et~al.(2024)Zhu, Liao, Zhang, Wang, Liu, and Wang]{pmlrv235zhu24f}
Lianghui Zhu, Bencheng Liao, Qian Zhang, Xinlong Wang, Wenyu Liu, and Xinggang
  Wang.
\newblock Vision mamba: Efficient visual representation learning with
  bidirectional state space model.
\newblock In \emph{Proceedings of the 41st International Conference on Machine
  Learning}, pages 62429--62442. PMLR, 2024.

\bibitem[Zhu et~al.(2010)Zhu, Singh, Siddiqui, and Gillam]{zhu2010automatic}
Yingxuan Zhu, PD Singh, Khan Siddiqui, and Michael Gillam.
\newblock An automatic system to detect and extract texts in medical images for
  de-identification.
\newblock In \emph{Medical Imaging 2010: Advanced PACS-based Imaging
  Informatics and Therapeutic Applications}, page 762803. SPIE, 2010.

\bibitem[Ziad et~al.(2016)Ziad, Alanwar, Alzantot, and
  Srivastava]{ziad2016cryptoimg}
M~Tarek~Ibn Ziad, Amr Alanwar, Moustafa Alzantot, and Mani Srivastava.
\newblock Cryptoimg: Privacy preserving processing over encrypted images.
\newblock In \emph{2016 IEEE Conference on Communications and Network Security
  (CNS)}, pages 570--575. IEEE, 2016.

\end{thebibliography}
}

\end{document}